\documentclass[11pt]{article}
\usepackage[preprint]{acl}

\usepackage{times}
\usepackage{latexsym}
\usepackage[T1]{fontenc}
\usepackage[utf8]{inputenc}
\usepackage{microtype}
\usepackage{inconsolata}
\usepackage{graphicx}

\usepackage{booktabs}
\usepackage{multirow}
\usepackage{amsmath}
\usepackage{enumitem}
\usepackage{xspace}
\usepackage{tikz}
\usetikzlibrary{positioning, arrows.meta, calc}
\DeclareRobustCommand{\cmark}{\tikz[baseline=-0.55ex]{\draw[green!55!black,line width=0.9pt] (0,0.6pt)--(2pt,-1.4pt)--(5pt,3.6pt);}}

\DeclareRobustCommand{\checkmark}{\cmark}
\newcommand{\fininteract}{\textsc{FinInteract}\xspace}
\newcommand{\interactcomp}{\textsc{InteractComp}\xspace}

\newcommand{\disep}{$\text{DisE}^+$\xspace}

\newcommand{\limref}{the Limitations section}

\newcommand{\ethicscode}{the ACL Code of Ethics}

\newcommand{\GPUSPEC}{a single NVIDIA H200 GPU}

\title{FinInteract: Benchmarking Clarification and Intent Integration in Ambiguous Financial Question Answering}

\author{
  \textbf{Xinyu Wang} \\
  McGill University
  \And
  \textbf{Tung Sum Thomas Kwok} \\
  University of California, Los Angeles
  \And
  \textbf{Zhenghan Tai} \\
  University of Toronto
  \AND
  \textbf{Guang Cheng} \\
  University of California, Los Angeles \\
  \texttt{guangcheng@stat.ucla.edu}
}

\begin{document}
\maketitle

\begin{abstract}
Large language model agents increasingly answer financial questions by searching regulatory filings. Such questions are often deceptively under-specified: Meta Platforms' ``operating income'' is \$46.75B consolidated but \$62.87B for the Family of Apps segment, and each reading is exactly verifiable against the filing. A capable agent should recognize the ambiguity and ask, rather than commit to a plausible but unintended reading. Existing financial benchmarks cannot measure this, because one gold answer per question cannot separate agents that resolve the ambiguity from those that guess the common reading, a blind spot we call the \emph{single-gold illusion}. We release \fininteract, a bilingual (English/Chinese) benchmark of 173 instances that pairs each question with a \emph{default} and an \emph{intended} interpretation across a five-category ambiguity taxonomy, and grades whether an agent elicits the right clarification and then integrates it. Re-grading identical outputs against the default rather than the intended reading inflates GPT-4o's accuracy by 3.1 times, confirming the illusion. Beyond it, we find that models answer above 90\% once the interpretation is supplied but at most 28.9\% when they must elicit it themselves, that targeting is uneven across a taxonomy well powered for entity scope and metric definition and exploratory elsewhere, and that conditioning on the ambiguity category improves resolution at both inference and training time.

\end{abstract}

\section{Introduction}
\label{sec:intro}
Large language model (LLM) agents are increasingly utilized in financial analysis to answer questions in regulatory filings~\citep{islam2023financebench,chen2021finqa,reddy2024docfinqa}, and carry out multi-step research and retrieval over corporate disclosures~\citep{bigeard2025financeagent,choi2025finagentbench,krumdick2024bizbench,xie2024finben}. The agent reads a user's question, searches through relevant filings, and returns the answer for downstream actions.

\begin{figure*}[t]
\centering
\includegraphics[width=0.8\textwidth]{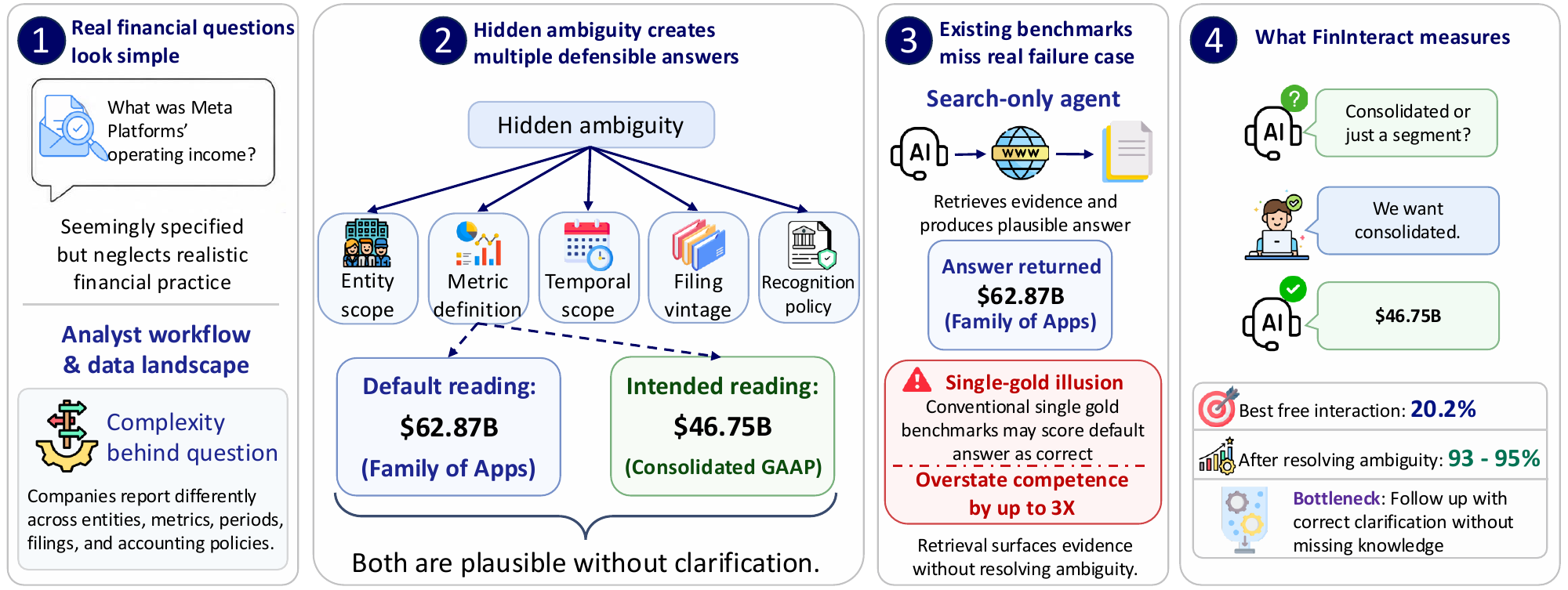}
\caption{\textbf{Evaluating financial agents on ambiguous queries.}
\textbf{(1-2)}~A real query (``What was Meta Platforms' operating income?'') looks fully
specified but hides ambiguity across five categories (entity, metric, period, filing, policy),
here giving two defensible answers: a default \$62.87B (Family of Apps segment) and the intended
\$46.75B (consolidated GAAP). \textbf{(3)}~A search-only agent returns the plausible default,
which a single-gold benchmark scores as correct, the \emph{single-gold illusion} that overstates
competence by up to 3 times. \textbf{(4)}~\fininteract instead grades whether the agent
\emph{asks}: the GPT-5 agent reaches 20.2\% under free interaction versus 93-95\% once the
ambiguity is resolved, so the gap lies in interaction, not missing knowledge.}
\label{fig:teaser}
\end{figure*}

However, financial questions are often deceptively under-specified. Financial terminology is highly specific, and a natural language question may map to several defensible readings for different numbers, e.g., in the question \emph{``What was Meta Platforms' operating income?''} (Figure~\ref{fig:teaser}), an agent must decide whether the user means the consolidated GAAP figure (\$46.75B for FY2023) or the Family of Apps segment (\$62.87B), both being programmatically verifiable against SEC facts. A capable agent should recognize the ambiguity and follow up with the user instead of directly committing to a plausible but unintended reading.

Existing financial benchmarks cannot measure this as they grade only the final answer by assuming the question is disambiguated~\citep{islam2023financebench,chen2021finqa,reddy2024docfinqa,krumdick2024bizbench,xie2024finben}. Without a grounding check, they cannot distinguish whether the agent actually resolves the ambiguity or guesses the common reading. We call this blind spot the \emph{single-gold illusion} (Table~\ref{tab:comparison} in Appendix~\ref{app:related} contrasts \fininteract with these benchmarks), which is currently underexplored in the financial domain relative to general ambiguous QA literature~\citep{min2020ambigqa,li2025condambigqa,kuhn2022clam}. We study financial ambiguity over three research questions: \begin{itemize}[leftmargin=*, topsep=3pt, itemsep=1pt, parsep=0pt]
    \item \textbf{RQ1}: Do frontier models recognize financial report ambiguity? 
    \item \textbf{RQ2}: Which financial ambiguity categories lead to the most failures?
    \item \textbf{RQ3}: Can interventions at inference or training time improve resolution?
\end{itemize}

Answering these questions requires four design goals, which we state here and instantiate in Sec.~\ref{sec:methodology}. \textbf{(G1) Representative coverage}: the benchmark must draw real ambiguity from public filings. \textbf{(G2) Fine-grained diagnosis}: it must expose the distinct dimensions of ambiguity through paired default and intended readings, so that evaluation can localize which clarification ability a model lacks. \textbf{(G3) Low-cost, leak-safe construction}: it must be buildable semi-automatically, and the disambiguating context must never contain the answer. \textbf{(G4) High, human-validated quality}: ground truth must be programmatically verifiable and validated by domain experts at minimal annotation cost.

We release \fininteract, a bilingual benchmark in English and Chinese containing 173 instances, classified into five financial ambiguity categories. Each question $Q$ is paired with a \emph{default} interpretation recording a possible non-expert assumption and an \emph{intended} interpretation fixed by the disambiguating context. By grading the same model outputs against the default and intended interpretations, we effectively expose this single-gold illusion. \fininteract includes a skill metric that localizes which clarification ability the model lacks, and evaluates the agents under a ReAct protocol where the user simulator answers only from the disambiguating context (Figure~\ref{fig:framework}) to measure the agent's quality of clarify-then-integrate action. 

Our evaluation answers the research questions in turn. First, while frontier models detect ambiguity, they rarely resolve it effectively: they ask the correct clarification question but still \textbf{fail to convert it into the correct answer}, reaching at most 28.9\% under their own clarification against a 93-95\% ceiling once the interpretation is supplied (Sec.~\ref{sec:ceiling}). Second, targeting is \textbf{uneven across ambiguity categories}: entity, metric, and temporal ambiguity are almost always named, while no model ever raises the recognition basis, though this last observation rests on an exploratory nine-instance category (Sec.~\ref{sec:skills}). Third, \textbf{conditioning on the ambiguity category helps at both inference and training time}: a category-aware clarification policy lifts GPT-5 from 34.0\% to 46.0\% on a stratified pilot, and category-guided fine-tuning raises on-category targeting from 25 to 69 AC@1, suggesting that this benchmark is not only diagnostic but also actionable (Sec.~\ref{sec:trainsignal} and Appendix~\ref{app:findings}). This paper makes three contributions: \begin{enumerate}[leftmargin=*,itemsep=1pt]
  \item \textbf{Problem definition.} We formalize ambiguity in financial questions with five categories in standard accounting and disclosure practice (Sec.~\ref{sec:task}).
  \item \textbf{Benchmark.} We introduce \fininteract, a bilingual financial ambiguity dataset of 173 data, containing a paired default and intended interpretations verified by domain experts (Sec.~\ref{sec:methodology}).
  \item \textbf{Empirical findings.} Our evaluation shows that existing models commonly fail to return the correct final answer even when they have correctly clarified the intention, and that conditioning on the ambiguity category improves resolution at both inference and training time (Sec.~\ref{sec:exp}).
\end{enumerate}

\section{Problem Formulation}
\label{sec:task}
We first define an ambiguous financial QA instance, then the five-category taxonomy, the entropy measure that scores instance difficulty, and the interaction protocol under which agents are evaluated. An instance (example in Table~\ref{tab:example}) consists of an ambiguous question $Q$ over a company filing and a disambiguating context $C$, the minimal set of scope or definitional constraints that collapses $Q$ into a unique intended answer $A$ under the intended interpretation $\mathcal{I}_{\text{int}}$. The same question also admits a second correct answer $A_{d}\neq A$ under the default interpretation $\mathcal{I}_{\text{def}}$, the reading a non-expert would assume when seeing $Q$ alone. Both $\mathcal{I}_{\text{int}}$ and $\mathcal{I}_{\text{def}}$ are structured records over the financial ambiguity taxonomy defined below.
\begin{table}[h]
\centering
\caption{This ambiguous financial question example admits three plausible entity scopes, i.e., the consolidated total and two reportable segments, giving an entropy of 1.58 bits. Without the disambiguating context $C$, a model's default interpretation yields the wrong answer.}
\label{tab:example}
\small
\begin{tabular}{p{0.30\columnwidth}p{0.55\columnwidth}}
\toprule
Field & Value \\
\midrule
$Q$ &
  What was Meta Platforms' operating income? \\
$C$ &
  Total company (consolidated) results under U.S.\ GAAP for the year
  ended December 31, 2023. \\
$A$ (intended) &
  \$46.75B \emph{[consolidated GAAP]} \\
$A_d$ (default) &
  \$62.87B \emph{[Family of Apps segment]} \\
\midrule
Ambiguity category & Entity scope \\
$H_0$ & 1.58 bits \\
\midrule
Intended span &
  ``Income from operations for 2023 was \$46.75 billion\ldots'' \\
Default span &
  ``Family of Apps\ldots Income from operations \$62,871\ldots'' \\
\bottomrule
\end{tabular}
\end{table}

\textbf{Financial ambiguity taxonomy.}
We derive five categories of financial ambiguity from standard
accounting and disclosure practice (Table~\ref{tab:axes}). Each instance normally exercises one primary category, with one or two optional secondary categories when additional ambiguity is present.

\begin{table}[h]
\centering
\caption{Five-category financial ambiguity taxonomy.}
\label{tab:axes}
\small
\begin{tabular}{p{0.23\columnwidth}p{0.63\columnwidth}}
\toprule
Category & Definition \\
\midrule
\textbf{Temporal scope} &
  Ambiguity over which time period is intended: fiscal year ending
  September vs. calendar year; Q3 results vs. full year; TTM vs.
  point-in-time. \\
\addlinespace
\textbf{Metric definition} &
  Ambiguity over accounting basis: GAAP net income vs. non-GAAP adjusted
  income; organic revenue vs. as-reported; operating EBITDA vs. net income. \\
\addlinespace
\textbf{Entity scope} &
  Ambiguity over consolidation level: segment vs. consolidated total;
  parent-attributable vs. full consolidated; Class A vs. Class B shares;
  A-shares vs. H-shares. \\
\addlinespace
\textbf{Filing vintage} &
  Ambiguity over filing version: original 10-K vs. amended 10-K/A;
  as-reported vs. restated; preliminary earnings release vs. final filing. \\
\addlinespace
\textbf{Recognition policy} &
  Ambiguity over accounting policy: revenue recognized point-in-time
  vs. over time; gross vs. net revenue; capitalized vs. expensed R\&D. \\
\bottomrule
\end{tabular}
\end{table}

\textbf{Disambiguation entropy.} We quantify how ambiguous an instance is using the prior confidence entropy $H_{0} = \log_{2} k$, where $k$ denotes the number of plausible interpretations of $Q$ without clarifying interaction, e.g., the example in Table~\ref{tab:example} has three plausible interpretations and hence an entropy of $H_0 = \log_2 3 \approx 1.58$ bits. $H_0$ also serves as our measure of difficulty and clarification efficiency in Sec.~\ref{sec:metrics}. Both $A$ and $A_{d}$ are defensible from publicly available filings, so an agent should be able to identify the ambiguity by reading $Q$ alone without $C$. This pairing of default and intended interpretations generalizes the target-distractor design of \interactcomp~\citep{interactcomp2026} to the financial domain.

\textbf{Agent interaction model.} We follow \interactcomp's ReAct framework~\citep{yao2023react} where the agent may issue three types of actions, namely \textbf{search}($Q$) to retrieve a passage from the filing corpus, \textbf{interact}($Q$) to present a yes-or-no question to the simulated user who answers based on $C$, and \textbf{answer}($Q$) to submit a final answer. Specifically, we evaluate agents in three primary modes (Table~\ref{tab:modes}) and four
ablation modes. We additionally include a category-conditioned interaction mode, described in
Appendix~\ref{sec:axisreact}.

\begin{table}[h]
\centering
\caption{We evaluate interaction modes including answer-only (A), answer with search (A+S) and answer with search and interaction (A+S+I).}
\label{tab:modes}
\small
\begin{tabular}{p{0.34\columnwidth}p{0.52\columnwidth}}
\toprule
Mode & Description \\
\midrule
A & Pure parametric recall \\
A+S & Oracle retrieval augmented \\
A+S+I & Standard full interaction \\
\midrule
Always-ask & $\geq 1$ follow-up question \\
Category-oracle & Provide primary category\\
Template-oracle & Fixed human-written question \\
Enumerate & Lists all interpretations and answers \\
\bottomrule
\end{tabular}
\end{table}

\section{Methodology}
\label{sec:methodology}
\begin{figure*}[t]
\centering
\includegraphics[width=0.8\textwidth]{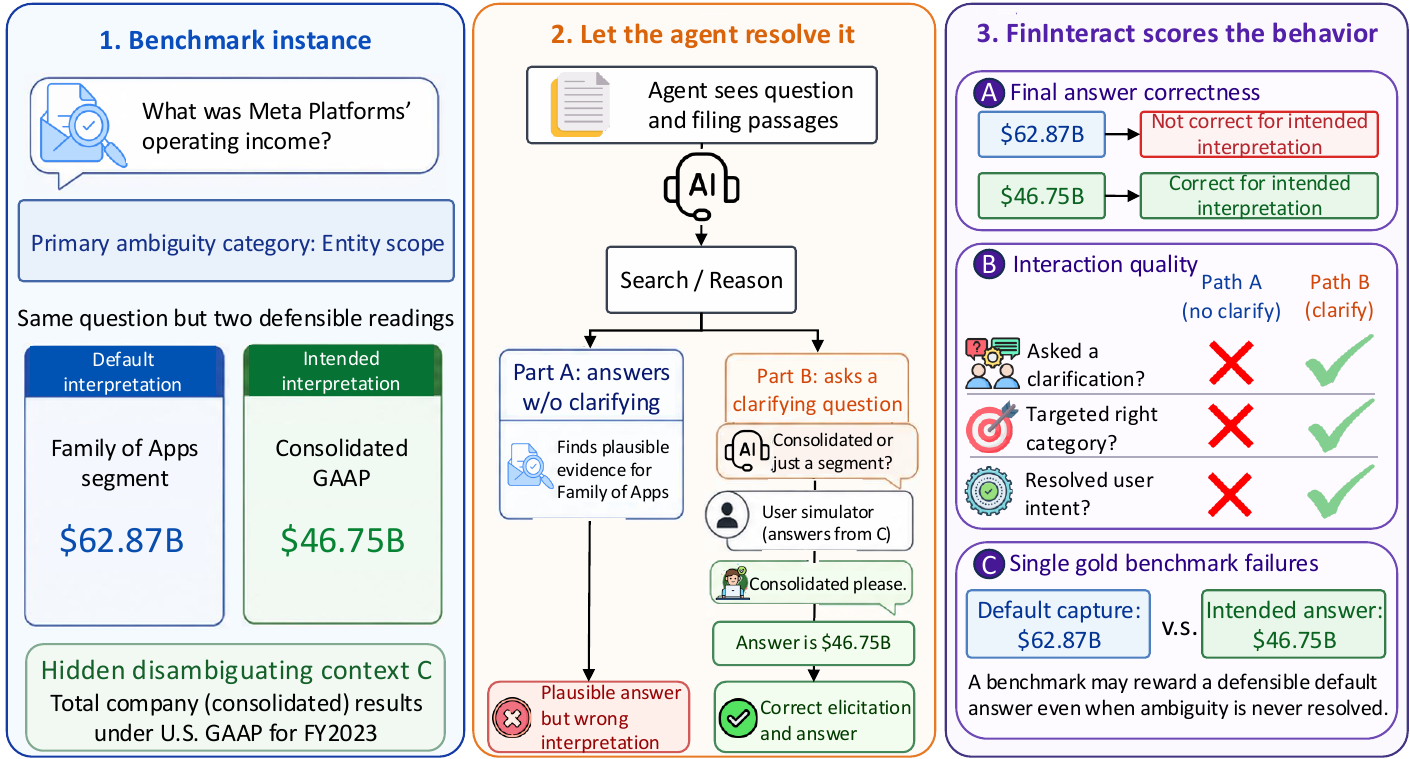}
\caption{\textbf{\fininteract rewards actions that resolve the intended interpretation.}
\textbf{(1)}~Each instance carries a primary ambiguity category and two defensible readings, a
default (Family of Apps, \$62.87B) and an intended (consolidated GAAP, \$46.75B), separated by a
hidden disambiguating context $C$. \textbf{(2)}~The agent either answers without clarifying
(Part~A, a plausible but wrong reading) or asks a question that a user simulator answers only from
$C$ (Part~B, correct elicitation). \textbf{(3)}~\fininteract scores on final-answer
correctness against the intended interpretation, and interaction quality (did it ask, target the right
category, and resolve intent).}
\label{fig:framework}
\end{figure*}

\fininteract instantiates the formulation of Sec.~\ref{sec:task} as a semi-automated ambiguous financial QA corpus built from public filings. Each instance carries a programmatically verifiable answer and a paired default and intended interpretation, produced by an LLM construction pipeline with minimal human annotation and validated by domain experts. Figure~\ref{fig:framework} shows a worked instance and how the benchmark scores an agent's two resolution paths.

\subsection{Design goals and principles}
\textbf{Design goals.} \fininteract targets four goals: \textbf{(G1) Representative coverage} to draw real ambiguity from public filings. 
\textbf{(G2) Fine-grained diagnosis} to contain diverse possible readings so that evaluation can localize which clarification ability a model lacks. \textbf{(G3) Low-cost, leak-safe construction} to build and verify the corpus at scale in a semi-automated setting. \textbf{(G4) High, human-validated quality} to validate the instances and judges by domain experts with minimal annotation cost.

\textbf{Design principles.} To satisfy the four goals, we build \fininteract under four principles. \textbf{(P1) Easy to verify, ambiguous to resolve}: ground truth is programmatically verifiable from SEC XBRL facts and CNINFO structured data, yet the question cannot be resolved correctly without the disambiguating context $C$. \textbf{(P2) No yes/no questions}: the agent must produce a grounded value rather than a binary judgment. \textbf{(P3) Company name included}: the entity itself is never ambiguous, using a capitalized proper noun in English and a CJK company name in Chinese. \textbf{(P4) Answer absent from $C$}: the agent cannot take the shortcut of copying directly from the context.

\subsection{Data sources}
\label{sec:stats}
We build the instances from official filings. English passages, which make up about 80\% of the candidate pool, come from SEC filings: DocFinQA~\citep{reddy2024docfinqa} long-context 10-K and 10-Q passages with verified QA pairs, and XBRL-derived facts for S\&P~500 and Russell~1000 companies for fiscal years 2024-2025. Chinese passages come from the A-share annual reports of SSE~50 and CSI~300 constituents, retrieved through the CNINFO and akshare structured APIs. We avoid reusing published QA pairs directly to reduce memorization risk, though we do not claim absolute contamination safety. The current evaluation corpus contains 173 instances over 61 companies drawn from 2022-2025 filings, labeled with the taxonomy of Sec.~\ref{sec:task}; Table~\ref{tab:stats} gives the detailed breakdown. Specific details regarding passage counts and difficulty distribution are included in Appendix~\ref{app:construction}.

\begin{table}[t]
\centering
\caption{FinInteract dataset statistics. Disambiguation entropy
$H_0=\log_2 k$ over $k$ plausible interpretations.}
\label{tab:stats}
\small
\setlength{\tabcolsep}{4pt}
\begin{tabular}{@{}lr@{\hskip 1.5em}lr@{}}
\toprule
Property & Value & Property & Value \\
\midrule
\textbf{Total instances}     & 173 & \textbf{Category}: entity   & 94 \\
\textbf{Language}: EN         & 53  & \quad metric        & 63 \\
\quad ZH             & 120 & \quad recognition   & 9 \\
\textbf{Source}: EDGAR        & 50  & \quad temporal      & 7 \\
\quad CSRC-CNINFO    & 120 & \quad filing        & 0 \\
\quad DocFinQA       & 3   & \textbf{\# cat.}: single     & 128 \\
\textbf{Filing}: 10-K         & 53  & \quad two           & 45 \\
\quad annual report  & 120 & \textbf{$H_0$}: mean         & 2.20 \\
\textbf{Filing yr}: 2022      & 40  & \quad median        & 2.00 \\
\quad 2023           & 53  & \quad range         & 1.0-3.6 \\
\quad 2024           & 58  & \textbf{Blind-solve rate}    & 0.7\% \\
\quad 2025           & 19  &  \textbf{Unique companies}     & 61  \\
\bottomrule
\end{tabular}
\end{table}

\subsection{Curation}
\label{sec:construction}
We curate each instance through a three-role pipeline. A \textbf{constructor} agent based on GPT-5 generates the full $(Q, C, A)$ triple, together with the default and intended interpretations and the category label, from a filing passage; category-specific instructions prevent drift toward the categories that are easiest to generate. Next, a code-level \textbf{sanity check} rejects malformed candidates immediately, without API calls. Finally, an \textbf{adversarial verifier} runs ten trials that answer $Q$ without the disambiguating context $C$, and rejects an instance if at least two of the ten produce the intended answer and the intended interpretation simultaneously. Details of the prompts, sanity-check rules, output schema, and cost are in Table~\ref{tab:qc}, Appendix~\ref{app:construction}, and Appendix~\ref{app:cost}.

\subsection{Validation}
\label{sec:humanval}
Two finance-literate annotators, holding the Chartered Financial Analyst (CFA) designation and a Hong Kong Securities and Futures Commission (SFC) license, independently review instances to check whether (1) the question $Q$ is ambiguous, (2) the disambiguating context $C$ uniquely identifies the intended reading without stating the answer, and (3) both the intended and the default answers are correct. A separate panel of five annotators supports the grader-validation and default-reading studies reported in Appendix~\ref{app:maindetails}. We record substantial agreement, averaging 0.85 Gwet's AC1~\citep{gwet2008}, indicating strong chance-corrected consensus that, unlike Cohen's $\kappa$, stays reliable under the heavy skew toward ``yes'' labels. The full protocol and annotation details are in Appendix~\ref{app:construction}.

\subsection{Metrics}
\label{sec:metrics}

We adopt standard metrics from the ambiguous QA and interactive retrieval literature~\citep{chen2021finqa,interactcomp2026,islam2023financebench,wei2025browsecomp} so that \fininteract can be compared directly to prior benchmarks.

\textbf{Accuracy (\%)} is the fraction of instances whose final answer matches the \emph{intended} interpretation, graded by GPT-4o-mini under a financial-domain tolerance covering measurement differences, entity equivalence across stock markets, and currency normalization, with a fiscal-year mismatch counted as wrong.

\textbf{Default-capture rate (\%)} is the fraction whose final answer instead matches the \emph{default} interpretation. It measures how often a model returns the naive reading, letting us reconstruct what a conventional single-gold benchmark would have scored (Sec.~\ref{sec:illusion}).

\textbf{Interaction diagnostics} comprise the interaction rate (IR), the proportion of instances on which the agent issues at least one clarification question; the average number of clarification turns (Round); and the accuracy of targeting the true ambiguity category (AC@1). We report AC@1 in two forms, according to whether the question names any true category and whether it makes the single most important category its central concern.

\textbf{Confidence} is measured by the Expected Calibration Error (ECE) between stated confidence and correctness, capturing an agent's overconfidence on ambiguous queries.

Full metric definitions, including the accuracy grading tolerance, the AC@1 decomposition, and a supplementary clarification-efficiency score (\disep), are in Appendix~\ref{app:metrics}.

\section{Experiments}
\label{sec:exp}
\label{sec:eval}
\subsection{Evaluation setup}
\textbf{Agent architecture.} We evaluate agents under the ReAct~\citep{yao2023react} framework in the seven configurations of Table~\ref{tab:modes}. The \textbf{interact} action presents a yes-or-no question to the simulated user, who must respond ``yes'', ``no'', or ``I don't know''.

\textbf{Prompting strategy.} \fininteract evaluates interactive agents under several interaction protocols: answer-only, answer with search, answer with search and interaction, and the category-conditioned policies. All models run under the same ReAct system prompt. Additional details are in Appendix~\ref{sec:error}.

\textbf{Models.} We evaluate a diverse set of models. The proprietary set comprises the GPT family (GPT-5.6, GPT-5, GPT-4o, and GPT-5-mini), Claude-Sonnet-5, Gemini-3.5-flash, Grok-4.5, DeepSeek-V4-flash, GLM-5.2, and Kimi-K3. The open-source set comprises the dense Qwen3 family (4B, 8B, 14B, and 32B) and two mixture-of-experts models, Qwen3-30B-A3B and Qwen3.5-35B-A3B. Proprietary models are accessed through a unified OpenRouter harness.

\subsection{Overall performance}
\label{sec:ceiling}
Even frontier models are vulnerable to financial report ambiguity (Table~\ref{tab:main}). No model exceeds 44\% accuracy on the intended interpretation under free interaction, and the ceiling is low for proprietary and open-source models alike: the best full-scale model reaches 28.9\% and the best cross-vendor pilot model 44.0\%. The opportunity to interact does not by itself improve accuracy, as the interaction difference $\Delta$ (added-interaction minus plain-search accuracy) is non-positive for eleven of the sixteen evaluated models. This suggests that the failure is not one of asking but of converting a clarified intention into the correct answer. This is supported by over 80\% AC@1 and an interaction rate well above the accuracy, indicating that agents do recognize which kind of ambiguity is present and do ask about it, yet still answer wrongly. 

Two ablations bracket this gap (Table~\ref{tab:ceiling}). Forcing four rounds of clarification \emph{lowers} accuracy rather than raising it, which is consistent with agents getting lost in multi-turn conversations~\citep{laban2026llms}. Supplying the disambiguating context $C$ as an oracle instead lifts every model to 93-95\%. Because $C$ never contains the final answer $A$, this improvement shows that the gap is not missing knowledge but the ability to resolve the interpretation and carry it into the answer. Appendix~\ref{app:maindetails} reports the full analysis. \textbf{Finding 1: models can observe the ambiguity and answer correctly once the intention is made explicit, but fail to bridge the intermediate step from an updated intention to the correct answer.}

\begin{table}[t]
\centering
\caption{Forcing more clarification does not help accuracy, but \emph{resolving} the ambiguity lifts every model to 93-95\%. The gap is therefore the model's inability to elicit $C$ for itself, not missing knowledge or grounding. Abbreviations follow Table~\ref{tab:modes}.}
\label{tab:ceiling}
\small
\begin{tabular}{lcccc}
\toprule
Model & A & A+I & Forced(4) & \textbf{Oracle} \\
\midrule
GPT-5      & 0.6 & 20.2 & 1.5 & \textbf{95.4} \\
GPT-4o     & 0.6 & 4.6  & 0.0 & \textbf{93.1} \\
GPT-5-mini & 0.6 & 0.0  & 0.6 & \textbf{95.4} \\
Qwen3-30B   & 1.2 & 11.6 & -- & \textbf{90.2} \\
Qwen3.5-35B & 0.6 & 28.9 & -- & \textbf{91.9} \\
\bottomrule
\end{tabular}
\end{table}

\newlength{\midgap}\setlength{\midgap}{0.02\textwidth}
\begin{table*}[t]
\centering
\caption{\textbf{(a)}~Main results, reporting accuracy (\%) against the \emph{intended}
interpretation per mode, with $\Delta$ the +Interact minus +Search difference.
\textbf{(b)}~Re-grading the same outputs against the intended versus the default reading a conventional benchmark would treat as gold (Same abbreviation as Table~\ref{tab:modes}). \textbf{(c)}~Per-category targeting
(AC@1), where recognition policy is a shared blind spot ($=0$ for every model). Category
sizes: entity (Ent.) 94, metric (Met.) 63, recognition (Rec.) 9, temporal (Tmp.) 7.}
\label{tab:main}
\begin{minipage}[t]{0.53\textwidth}
\centering
\textbf{(a) Main results}\par\smallskip
{\small
\resizebox{\columnwidth}{!}{%
\begin{tabular}{lccccccc}
\toprule
& \multicolumn{2}{c}{Acc (\%)} &
  \multicolumn{5}{c}{Answer+Search+Interact} \\
\cmidrule(lr){2-3}\cmidrule(lr){4-8}
Model & A & A+S & A+S & $\Delta$ & IR & AC@1 & ECE \\
\midrule
\multicolumn{8}{l}{\textit{Proprietary models}} \\
GPT-5.6           & 10.0 & 36.0 & 30.0 & $-6$  & -- & .98 & -- \\
GPT-5             & 0.6 & 6.4  & \textbf{20.2} & $+13.8$ & 98.8 & .86 & .53 \\
GPT-4o            & 0.6 & 12.1 & 4.6  & $-7.5$  & 99.4 & .94 & .82 \\
GPT-5-mini        & 0.6 & 0.0  & 0.0  & $0.0$   & 96.0 & .94 & .00 \\
Claude-Sonnet-5   & 4.0  & \textbf{46.0} & \textbf{44.0} & $-2$  & -- & .98 & -- \\
Grok-4.5          & 2.0  & 40.0 & 32.0 & $-8$  & -- & .87 & -- \\
GLM-5.2           & 12.0 & 40.0 & 40.0 & $0$   & -- & .96 & -- \\
Kimi-K3           & 16.0 & 38.0 & 34.0 & $-4$  & -- & .98 & -- \\
DeepSeek-V4-flash & 12.0 & 28.0 & 24.0 & $-4$  & -- & .96 & -- \\
Gemini-3.5-flash  & 14.0 & 6.0  & 30.0 & $+24$ & -- & .96 & -- \\
\midrule
\multicolumn{8}{l}{\textit{Open source models}} \\
Qwen3-4B          & 0.0 & 8.1  & 12.1 & $+4.0$  & 28 & .85 & -- \\
Qwen3-8B          & 0.6 & 14.4 & 24.9 & $+10.5$ & 15 & .88 & -- \\
Qwen3-14B         & 0.0 & 18.5 & 22.5 & $+4.0$  & 36 & .74 & -- \\
Qwen3-32B         & 1.2 & 27.2 & 8.4  & $-18.8$ & 61 & .96 & -- \\
Qwen3-30B-A3B     & 1.2 & 28.3 & 11.6 & $-16.7$ & 72 & .90 & -- \\
Qwen3.5-35B-A3B   & 0.6 & \textbf{35.8} & 28.9 & $-6.9$ & 36 & .81 & -- \\
\midrule
\multicolumn{8}{l}{\textit{Human baseline ($N{=}50$)}} \\
Human             & -- & -- & \textbf{30.0} & -- & 100 & .72 & -- \\
\bottomrule
\end{tabular}}}
\end{minipage}\hspace{\midgap}%
\begin{minipage}[t]{0.4\textwidth}
\centering
\textbf{(b) Single-gold illusion}\par\smallskip
{\small
\resizebox{\columnwidth}{!}{%
\begin{tabular}{p{0.7cm}lccc}
\toprule
Model & Mode & Intended & Default & $\Delta$ \\
      &      & (true Acc) & (single-gold) & \\
\midrule
\multirow{3}{1cm}{GPT-5}  & A   & 0.6  & 2.3  & +1.7 \\
       & A+S & 6.4  & 10.4 & +4.0 \\
       & A+S+I & \textbf{20.2} & 12.1 & $-8.1$ \\
\midrule
\multirow{3}{1cm}{GPT-4o} & A & 0.6  & 0.6  & 0.0 \\
       & A+S & 12.1 & \textbf{37.0} & \textbf{+24.9} \\
       & A+S+I & 4.6  & 6.4  & +1.8 \\
\midrule
\multirow{3}{1cm}{GPT-5-mini} & A & 0.6 & 0.0 & $-0.6$ \\
           & A+S & 0.0 & 0.0 & 0.0 \\
           & A+S+I & 0.0 & 0.0 & 0.0 \\
\bottomrule
\end{tabular}}}
\par\bigskip
\textbf{(c) Per-category targeting (AC@1)}\par\smallskip
{\small
\resizebox{0.7\columnwidth}{!}{%
\begin{tabular}{lcccc}
\toprule
Model & Ent. & Met. & Rec. & Tmp. \\
\midrule
GPT-5    & .89 & .92 & \textbf{.00} & 1.00 \\
GPT-4o   & 1.00 & .97 & \textbf{.00} & 1.00 \\
GPT-5-m  & .99 & 1.00 & \textbf{.00} & 1.00 \\
Q3-30B   & .93 & .96 & \textbf{.00} & 1.00 \\
Q3.5-35B & .85 & .87 & \textbf{.00} & 1.00 \\
\bottomrule
\end{tabular}}}
\end{minipage}
\end{table*}

\subsection{Failure of existing benchmarks}
\label{sec:illusion}

Because every instance carries both a default and an intended answer, we can re-grade the \emph{identical} model outputs under the two conventions and read off what a conventional single-gold benchmark would have reported (Table~\ref{tab:main}(b)). In the retrieval-augmented regime where such benchmarks operate, grading against the default inflates the score sharply: GPT-4o is credited 37.0\% against the default but only 12.1\% against the intended reading, so a single-gold benchmark would have reported 3.1 times its true competence. The inflation reverses only under interaction, where GPT-5 scores 20.2\% intended against 12.1\% default, which is precisely the point: asking is what converts a default-anchored guess into the intended answer, and it is the one behavior single-gold grading cannot see. Appendix~\ref{app:maindetails} reports the full re-grading together with a human study confirming that annotators shown only $Q$ and the passage record the default reading as gold 66\% of the time. \textbf{Finding 2: a single-gold convention overstates competence by up to 3.1 times, because one gold answer per question encodes the default reading rather than the intended one.}

\subsection{Targeting varies by ambiguity category}
\label{sec:skills}
Aggregate AC@1 hides where clarification actually succeeds, so we decompose targeting by ambiguity category (Table~\ref{tab:main}(c)). Models name the correct category almost always for entity scope, metric definition, and temporal scope, but never for recognition policy, where AC@1 is 0 for every model evaluated. A case study makes the missed distinction concrete. Asked ``What were General Electric's revenues?'', a model may answer \$67.95B or \$26.79B depending on the recognition basis: the former is GAAP revenue combining over-time service recognition with point-in-time product recognition, the latter counts only revenue recognized at a point in time. Models return a single figure without asking which basis is meant, defaulting to the assumption that ``revenue'' is well defined and passing over the ASC~606 distinction between point-in-time and over-time recognition. We read this as evidence that a domain-specific reporting practice is not reflected in current models' general clarification behavior, which is what motivates the category-conditioned interventions of Sec.~\ref{sec:trainsignal}.

This last observation carries an important caveat. Recognition policy is one of our three exploratory categories, with only n = 9 instances, and an item-level audit (Appendix~\ref{app:limitations}) finds most of them to be ASC~606 timing disaggregations where neither value is the total a non-expert would assume. Our two human annotators also never targeted the category, so the blind spot is at least not model-specific. \textbf{Finding 3: targeting accuracy is uneven across ambiguity categories; entity, metric, and temporal ambiguity are reliably named, while recognition policy is never raised, on preliminary evidence from a nine-instance category.}

\subsection{Ambiguity category as training signal}
\label{sec:trainsignal}
\begin{figure*}[t]
\centering
\includegraphics[width=0.8\textwidth]{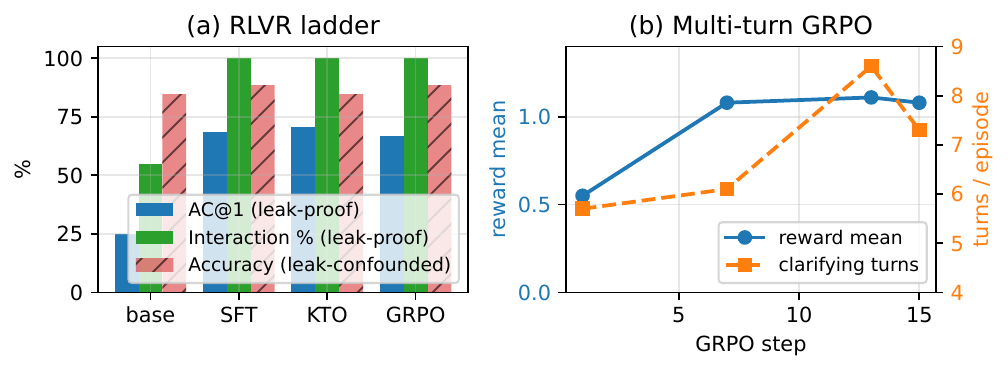} \vspace{-2mm}
\caption{\textbf{Category-guided training installs on-category clarification.} (a) The leak-proof signals, AC@1 and interaction rate, jump at the supervised stage and change little thereafter. (b) True multi-turn GRPO on the 4B policy: mean reward and clarifying turns rise together over training.}
\label{fig:grpo}
\end{figure*}

The diagnosis so far is descriptive, so we ask whether the ambiguity category can also serve as a training signal. We run category-conditioned training on a Qwen3-4B policy, supervised fine-tuning on category-guided demonstrations followed by reinforcement learning, teaching the model to interact and to target the correct category. On the held-out split the leak-proof signals rise almost entirely at the supervised stage, with AC@1 going from 25 to 69 and interaction from 55 to 100 (Table~\ref{tab:grpo}, Figure~\ref{fig:grpo}a), while the later stages move within run-to-run noise at n = 51. Full multi-turn reinforcement learning confirms that the episode is trainable end to end, with clarifying turns rising from 5.7 to 8.6 (Figure~\ref{fig:grpo}b). At inference time the same idea, conditioning on the category, lifts GPT-5 from 34.0\% to 46.0\% on a stratified N = 50 pilot (Appendix~\ref{sec:axisreact}). The gains concentrate on the data-abundant categories, so the signal is actionable but does not reach the recognition-policy blind spot of Sec.~\ref{sec:skills}. Additional details are in Appendix~\ref{sec:grpo}. \textbf{Finding 4: conditioning on the ambiguity category improves clarification at both training and inference time, so \fininteract provides an actionable signal and not only a diagnosis.}

\begin{table}[t]
\centering
\caption{RLVR ladder on \fininteract with Qwen3-4B, held-out n = 51. Category-guided supervision installs on-category clarification, with the leak-proof signals (AC@1 and interaction) rising almost entirely at the SFT stage. Accuracy is leak-confounded and is reported for completeness only.}
\label{tab:grpo}
\small
\begin{tabular}{lcccc}
\toprule
Stage & AC@1 & Interaction & Accuracy & Reward \\
\midrule
Base      & 25.0 & 54.9 & 84.3 & 0.96 \\
\;\;+\,SFT & 68.6 & \textbf{100} & 88.2 & 1.28 \\
\;\;+\,KTO               & \textbf{70.6} & \textbf{100} & 84.3 & 1.27 \\
\;\;+\,GRPO              & 66.7 & \textbf{100} & 88.2 & \textbf{1.30} \\
\bottomrule
\end{tabular}
\end{table}

\section{Discussion}
\label{sec:discussion}
We draw four takeaways here and develop each in Appendix~\ref{app:discussion}. \textbf{Ambiguity categories are different learning problems}, so an aggregate interaction score hides where the difficulty actually sits. \textbf{Elicitation is a trainable skill that current models lack}, so progress depends less on larger backbones than on objectives that reward asking about the right category at the right time. \textbf{Ambiguity persists in human-AI collaboration}, since models confidently return the default reading even with domain experts in the loop. \textbf{The paired construction should transfer to further categories, languages, and high-stakes domains}, wherever a single gold answer hides genuine ambiguity.

\section{Related Work}
\label{sec:related}
We include most relevant works to \fininteract and defer further related work in Appendix~\ref{app:related}.

\textbf{Financial QA benchmarks} such as FinanceBench~\citep{islam2023financebench}, FinQA~\citep{chen2021finqa}, and DocFinQA~\citep{reddy2024docfinqa}, together with recent agentic successors, enforce single-gold, single-turn evaluation and test numerical reasoning or retrieval without exploring whether a query is under-specified. To our knowledge, \fininteract is the first to make query ambiguity the central focus through paired default/intended interpretations and multi-turn interaction in the financial domain. We include comparison with existing financial QA benchmarks in Appendix~\ref{app:related}.

\textbf{Ambiguous QA and clarifying questions.} In the general domain, AmbigQA~\citep{min2020ambigqa}, CLAMBER~\citep{zhang2024clamber}, and QuestBench~\citep{li2025questbench} study whether a model identifies under-specification and asks the minimal clarifying question, building on a long clarifying-question line in retrieval and dialogue. \interactcomp~\citep{interactcomp2026} is our direct methodological ancestor, which we extend to finance with a domain-specific ambiguity taxonomy, per-category targeting (AC@1), and programmatically verifiable XBRL-grounded evidence pairs.

\textbf{Multi-turn interaction and selective answering.} Large models often \emph{lose track} over multi-turn conversations~\citep{laban2026llms}, and calibrated systems frequently do better by answering \emph{selectively} or abstaining under ambiguity~\citep{cole2023selectively,kuhn2022clam}. Our findings echo this in the financial setting, where forcing extra clarification degrades accuracy, but \fininteract turns the observation into a controlled, per-category diagnosis of \emph{which} clarification an agent should issue and \emph{whether} issuing it resolves the query, instead of only measuring whether more turns help.

\section{Conclusion}
\label{sec:conclusion}
We introduced \fininteract, to our knowledge the first bilingual benchmark that evaluates financial search agents on genuinely ambiguous queries. Its five-category ambiguity taxonomy, paired default and intended interpretation design, and construction pipeline with an adversarial verifier yield 173 instances that resist trivial disambiguation. Our central finding is that the bottleneck in handling financial QA ambiguity is interaction capability rather than financial knowledge, and that this capability has two parts: asking for the right clarification, and updating the answer once the clarification arrives. When the ambiguity is resolved for them, agents answer at 93-95\% accuracy, against at most 28.9\% when they must elicit the resolution themselves. This extends the interaction stagnation identified in \interactcomp to a domain whose terminology is far more specific, and where the cost of a confident wrong answer is correspondingly higher. We further turn the diagnosis into an intervention in Sec.~\ref{sec:trainsignal}, showing that \fininteract also serves as an informative training signal, and we release it with full construction, evaluation, and training code for reproducible follow-on work.

\section*{Limitations}
We note the following limitations of this study.
\begin{enumerate}[leftmargin=*,itemsep=1pt,topsep=3pt]
  \item \textbf{Scale.} A 173-instance corpus is small, a consequence of prioritizing per-instance construction rigor.
  \item \textbf{Language balance.} The English-to-Chinese split is 53:120, which may favor models pre-trained more heavily in one of the two languages.
  \item \textbf{Dominating categories.} Entity scope (n = 94) and metric definition (n = 63) dominate, against only n = 9 for recognition policy, which suggests that current models are not able to generate instances of the rarer categories in the first place.
  \item \textbf{Generator and evaluator bias.} Under API budget constraints we use the GPT-5 family for construction and simulation, which may bias results toward that family.
  \item \textbf{Data contamination.} Instances are freshly constructed from FY2022-2025 facts, but we cannot rule out that recently released models were pre-trained on those filings.
\end{enumerate}
Appendix~\ref{app:limitations} expands each of these.

\bibliography{refs}

\appendix
\section{Metric Details}
\label{app:metrics}

This appendix expands the metrics defined in \S\ref{sec:metrics}.

\paragraph{Accuracy grading.}
Accuracy is the primary ranking metric, mirroring \interactcomp, BrowseComp, and FinanceBench. A
GPT-4o-mini grader marks a final answer correct against the intended interpretation under
financial-domain tolerance: a plus or minus 1\% numeric tolerance, entity/ticker equivalence, and currency
normalization, with a fiscal-year mismatch counted as wrong.

\paragraph{Interaction targeting (AC@1).}
For each interact action, a GPT-4o-mini classifier decides whether the clarifying question targets a
true ambiguity category for that instance. We report the targeting accuracy AC@1 alongside the
any-category, wrong-category, and generic-ask rates. Because a single clarification is often compound (one question can name an entity, a
period, and a metric at once), AC@1 has two forms: a lenient \emph{touch} form (the question names
any true category) and a strict \emph{central} form (the single most-central category is a true
category). The touch form is validated against human annotators (\S\ref{sec:humanval}). These
diagnostics characterize \emph{how} a model interacts and do not rank models.

\paragraph{Calibration.}
The Expected Calibration Error (ECE) between stated confidence and correctness uses five equal-width
confidence bins.

\paragraph{Clarification efficiency (\disep).}
As a single supplementary number folding accuracy and clarification cost together, we report the
efficiency of \emph{successful} clarification,
\[
  \text{DisE}^+ = \mathbf{1}[\text{correct}] \times \mathbf{1}[n_\text{asks}{>}0] \times
  \frac{H_0}{n_\text{asks}},
\]
which is positive only when the agent asked at least one question and then answered correctly, and
decreases with redundant asks. We restrict it to the asked-and-correct case to avoid the perverse
incentive of an all-instances variant, which would award its maximum to a confident \emph{zero-ask}
answer, rewarding the very behavior the benchmark penalizes. \disep is a diagnostic of
clarification efficiency, never the ranking metric.

\section{Detailed Results for the Core Findings}
\label{app:maindetails}

This appendix gives the full tables, figures, and findings summarized in \S\ref{sec:illusion},
\S\ref{sec:ceiling}, and \S\ref{sec:skills}.

\subsection{The Single-Gold Illusion}
Before analyzing model behavior, we establish the core motivation empirically:
\emph{a conventional single-gold financial-QA benchmark cannot see the failure
\fininteract measures.} Existing benchmarks (FinanceBench, FinQA, DocFinQA)
attach one gold answer per question. Built from the same filings, such a
benchmark would almost always adopt the \emph{default} reading, the headline
figure a non-expert assumes, as that gold. Our paired construction lets us
quantify the consequence directly: every instance carries both an intended and a
default answer, so we re-grade the \emph{identical} model outputs under both
conventions (Table~\ref{tab:main}(b)).

\textbf{Finding 5: A single-gold convention over-states competence by up to
3 times, and the gap is the benchmark's reason to exist.} In the
retrieval-augmented regime where current single-gold benchmarks operate, the
larger models return the \emph{default} reading more often than the intended one:
GPT-4o is graded 37.0\% correct against the default but only 12.1\% against
the intended answer, a single-gold benchmark would report 3.1 times its true
competence. The model is not reasoning to the right answer; it is confidently
producing a well-grounded answer to the \emph{wrong reading of the question},
which a single-gold benchmark rewards and therefore cannot detect. (GPT-5-mini is
the instructive exception: it captures \emph{neither} reading because it asks
rather than commits, so single-gold grading does not flatter it, but
nor does it answer.) Critically,
interaction is the \emph{only} mode where intended accuracy exceeds
default-capture (GPT-5: 20.2 vs.\ 12.1): asking is what converts a
default-anchored guess into the correct intended answer. This is precisely the
capability single-gold benchmarks are structurally blind to, and what
\fininteract is built to measure.

\textbf{Human validation of the default.} The single-gold argument assumes a conventional
benchmark builder, seeing only the question and passage, would record the \emph{default}
reading as gold. We test this directly: five annotators, shown $Q$ and the passage but not
the resolving context $C$, independently record the answer they would treat as gold. They
converge strongly (93\% of items unanimous; mean inter-annotator agreement 0.98) and
select the \emph{default} reading 66\% of the time versus the intended reading 34\%, with
fewer than 8\% reporting that they noticed a second defensible answer. A single-gold
benchmark built from these filings would therefore encode the default, empirically confirming
the illusion's premise rather than assuming it.

\subsection{Latent Capacity and the Oracle Ceiling}
\begin{figure}[t]
\centering
\includegraphics[width=\linewidth]{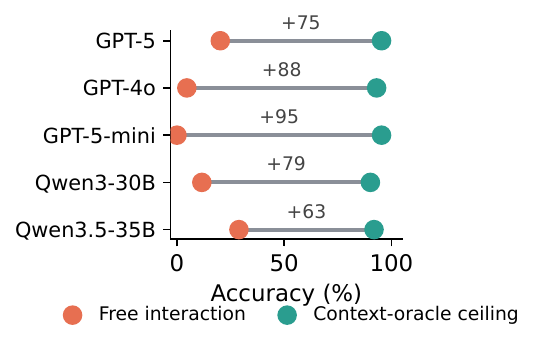}
\caption{\textbf{The elicitation gap.} Free-interaction accuracy (orange) versus the
context-oracle ceiling (teal) per model. Once the ambiguity is resolved every model
answers 90--95\% correctly, so the 63--95 point gap is the cost of the model
failing to \emph{elicit} the disambiguation itself, not missing knowledge.}
\label{fig:bottleneck}
\end{figure}

To locate the bottleneck we bracket interactive accuracy with two ablations
(Table~\ref{tab:ceiling} and Figure~\ref{fig:bottleneck}). \textbf{Forced interaction} requires the agent to ask
n = 4 clarifying questions before answering. \textbf{Context-oracle} hands the
model the disambiguating context $C$ (which fixes the intended interpretation)
together with a passage containing \emph{both} the intended and the default
evidence spans, then asks it to answer, the ambiguity is pre-resolved but the
model must still ground the correct value.

\textbf{Finding 6: Models can answer once the ambiguity is resolved, they
cannot resolve it themselves.} The context-oracle ceiling is 93--95\% for all
three models, including GPT-5-mini, which scores 0\% under free interaction.
Latent answering capacity is therefore near-ceiling and roughly scale-invariant;
what varies, and what \fininteract measures, is whether a model can \emph{resolve}
the ambiguity end to end, eliciting the disambiguating context and then integrating
it into a committed answer. Crucially, \emph{forcing} clarification
does not recover this capacity: requiring four asks leaves GPT-4o and GPT-5-mini
near 0\% (0.0\% and 0.6\%) and \emph{lowers} GPT-5 from 20.2\% to 1.5\%, because forced asks
consume the round budget on low-value questions without converging on a committed
answer. This already points past pure elicitation: models frequently ask on-category
yet still fail to commit the right value (the dominant error mode, \S\ref{sec:error}),
so the bottleneck is targeted, self-initiated disambiguation \emph{and} the integration
of the resulting answer, which neither scale nor brute-force asking supplies.

\subsection{Per-Category Skills}
\begin{figure}[t]
\centering
\includegraphics[width=\linewidth]{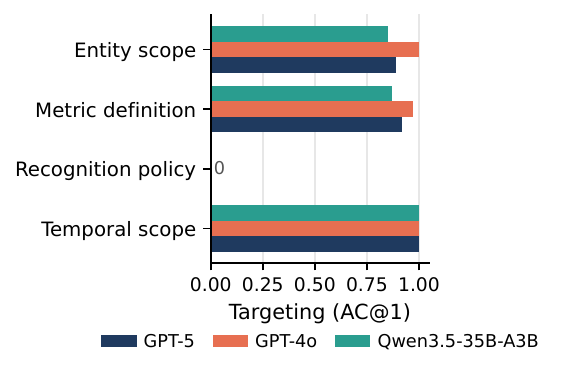}
\caption{\textbf{Per-category clarification-capability profile} (touch AC@1 from
Table~\ref{tab:main}(c)). Every model targets entity, metric, and temporal ambiguity well (0.85--1.00)
but collapses to zero on \texttt{recognition\_policy}, a blind spot shared across the
closed frontier and the strongest open model.}
\label{fig:radar}
\end{figure}

The five categories are not just data strata; each names a distinct \emph{skill}, the
ability to recognize that a question is under-specified \emph{along that category} and
ask the question that resolves it. We make this measurable by decomposing each
model's per-category \texttt{+interact} competence into three components:
\textbf{Recognition} (interaction rate, does it ask when it should?),
\textbf{Targeting} (AC@1, does it ask about the \emph{right} category?), and
\textbf{Resolution} (accuracy, does asking yield the intended answer?). A model
``has'' a category skill only when all three are high; a low value localizes the
failure to not-asking, asking-wrong, or asking-right-but-still-wrong.
Table~\ref{tab:main}(c) reports Targeting, the decisive component, and
Figure~\ref{fig:radar} shows the same profile as a per-category bar chart.

\begin{figure*}[t]
\centering
\includegraphics[width=\textwidth]{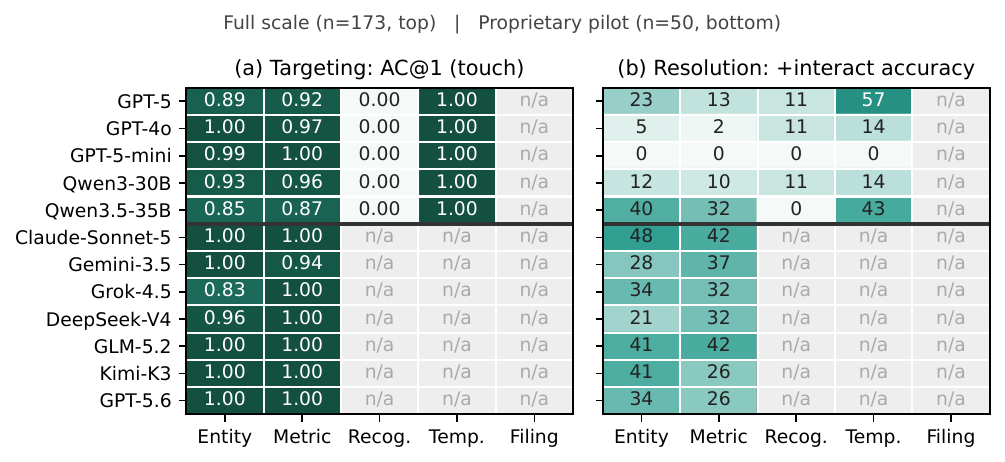}
\caption{\textbf{Per-model, per-category performance.} \emph{(a)}~Targeting (AC@1
touch) versus \emph{(b)}~Resolution (\texttt{+interact} accuracy) for every model and
ambiguity category. The top five rows are the full-scale panel (n = 173, all categories) and
the bottom seven are the proprietary pilot (n = 50), for which only entity and metric
carry enough instances (n = 29 and 19), with recognition, temporal, and filing at
$n$ at most 1 and shown n/a. \texttt{filing\_vintage} is empty at full scale too (n = 0).
Every model targets entity and metric almost perfectly, the proprietary models resolve
them better than the older OpenAI ladder, and no model targets recognition policy.}
\label{fig:axisperf}
\end{figure*}

\textbf{Finding 7: Ambiguity clarification is a per-category skill with a sharp touch
versus central-targeting dissociation.} Under the corrected classifier the pattern is
precise. \texttt{recognition\_policy} is a robust blind spot. No model ever asks about it,
in either the touch or the central form (AC@1 of 0), so models essentially never
question whether a figure is recognized point-in-time or over time. \texttt{temporal\_scope}
is fully solved. Models both touch it and make it the central concern (central AC@1
of 1.0). Entity and metric occupy a revealing middle. Models almost always \emph{touch} them
(touch AC@1 0.85--1.0) yet rarely make them the \emph{central} point of the question
(central AC@1 at most 0.16), defaulting instead to period confirmation. For these two categories
the failure is not neglect but imprecision, the model names the entity or metric in passing
while centering the question elsewhere. Resolution follows the same shape across all five models
(GPT-5 resolves 57\% of temporal but at most 23\% of entity and metric,
Figure~\ref{fig:axisperf}). The gap is not uniform incompetence
but missing specific targeting, which is precisely the intervention target
(\S\ref{sec:grpo}).

\textbf{Per-model, per-category case analysis.} Placing targeting and resolution side by
side for every model and category (Figure~\ref{fig:axisperf}) surfaces three patterns.
First, resolution does \emph{not} track targeting: GPT-4o and GPT-5-mini touch the
correct category on entity and metric almost as often as any model (AC@1 0.94--1.0)
yet resolve those categories at 0--5\%, so a high AC@1 is no guarantee of a correct
answer. Second, the model ranking \emph{reorders} across categories. Among the full-scale
models Qwen3.5-35B-A3B leads on the two abundant categories (entity 40\%, metric 32\%)
while GPT-5 leads on temporal (57\%). Third, the seven proprietary models resolve
entity and metric substantially better than the older OpenAI ladder (entity
21--48\%, metric 26--42\%, versus GPT-5's 23 and 13\%), yet they inherit the
\emph{same} recognition blind spot, with touch AC@1 0 on every recognition instance
they saw. \texttt{filing\_vintage} is untestable everywhere (n = 0). A single
aggregate interaction score therefore conceals which category a given model can actually
resolve, which is why we report per category.

\section{Additional Findings and Analyses}
\label{app:findings}

The main text reports four headline findings: the elicitation--integration gap, the single-gold illusion, the uneven per-category targeting, and the actionability of the ambiguity category as a training and inference signal. This appendix collects the remaining empirical findings and the full detail of the category-conditioned interventions.

\subsection{The Illusion Is Not Finance-Specific: A General-Domain Replication}
\label{sec:general}

\begin{table}[t]
\centering
\caption{The single-gold illusion replicates on AmbigQA (general domain,
n = 200, three models, \emph{same} grading methodology; the grader retries with
backoff so no answer is silently dropped). \emph{Intended}: accuracy against a
specified target reading. \emph{Any-valid}: accuracy against \emph{any} accepted
reading, what a lenient single-gold benchmark credits. \emph{Def.-cap}: fraction
producing a valid but non-target reading. \emph{Ask}: fraction that asks when
permitted. gpt-5-mini interact omitted (it asks below 1\%).
Modes: \emph{A} answer-only, \emph{A+O} answer with context oracle,
\emph{A+I} answer with interaction.}
\label{tab:general}
\small
\begin{tabular}{llcccc}
\toprule
Model & Mode & Intended & Any-valid & Def.-cap & Ask \\
\midrule
\multirow{3}{*}{\shortstack[l]{gpt\\4o-mini}}
 & A   & 31.5 & 72.5 & 55.0 & 0 \\
 & A+O & 49.5 & 68.5 & 31.0 & 0 \\
 & A+I & 43.5 & 81.0 & 53.0 & 27.0 \\
\midrule
\multirow{3}{*}{\shortstack[l]{gpt\\4o}}
 & A   & 39.0 & 84.0 & 65.5 & 0 \\
 & A+O & 68.5 & 81.5 & 31.5 & 0 \\
 & A+I & 43.0 & 86.0 & 64.0 & 2.5 \\
\midrule
\multirow{2}{*}{\shortstack[l]{gpt\\5-mini}}
 & A   & 36.0 & 77.0 & 62.5 & 0 \\
 & A+O & 62.0 & 73.5 & 27.5 & 0 \\
\bottomrule
\end{tabular}
\end{table}

\textbf{Finding 8: the single-gold illusion is domain-general.} Running the
identical default-vs-intended measurement on AmbigQA~\citep{min2020ambigqa}
(Table~\ref{tab:general}) reproduces all three effects outside finance,
\emph{consistently across three models}. Grading the \emph{same} answer-only outputs
against any accepted reading rather than a specified target overstates competence by
2.1--2.3 times (gpt-4o-mini 2.3, gpt-4o 2.2, gpt-5-mini 2.1), and the effect
is robust to how the target reading is chosen, a random-target assignment gives
2.1 times, versus our most-specific-reading default. Models return a valid but
non-target reading 55--66\% of the time, direct evidence the error is
ambiguity-blindness, not noise. Elicitation is again underused: even permitted to ask,
models clarify on at most 27\% of ambiguous queries and usually far less (gpt-4o
2.5\%, gpt-5-mini below 1\%), and interaction leaves intended accuracy far below the
any-valid ceiling (gpt-4o-mini 43.5 vs.\ 81). The one honest difference is the
oracle ceiling, 50--69\% here versus 93--95\% in finance, because
open-domain factoids stay hard even once disambiguated, so general knowledge is itself
a limiter. This is exactly why finance is the sharper instrument: its exact,
injectable ground truth isolates \emph{elicitation} from \emph{knowledge} in a way
general-domain QA cannot. The illusion itself, however, is a property of single-gold
grading, not of finance.

\subsection{Findings}

We report results for the full OpenAI tier ladder, GPT-5, GPT-4o, and
GPT-5-mini, across all three modes (Table~\ref{tab:main}). To probe whether the
effect is an OpenAI-specific artifact, we additionally evaluate a preliminary
seven-model, six-vendor panel of current frontier models (Claude-Sonnet-5,
Gemini-3.5-flash, Grok-4.5, DeepSeek-V4, GLM-5.2, Kimi-K3, GPT-5.6) through a
unified OpenRouter harness with an identical simulator/grader pipeline
(N = 50/mode, Table~\ref{tab:main}, panel (b)); it reproduces the headline pattern:
added interaction is flat-or-negative for six of seven vendors and yields a clear
gain only for Gemini, the weakest autonomous searcher. All headline accuracy
comparisons below are reported with
bootstrap 95\% CIs over instances, and we test each within-model mode
transition with a two-sided paired bootstrap (n = 173). Six findings emerge
here (Findings 9--14); the closely related latent-capacity ceiling is reported
as a headline result in the main text (\S\ref{sec:ceiling}).

\begin{figure}[t]
\centering
\includegraphics[width=\linewidth]{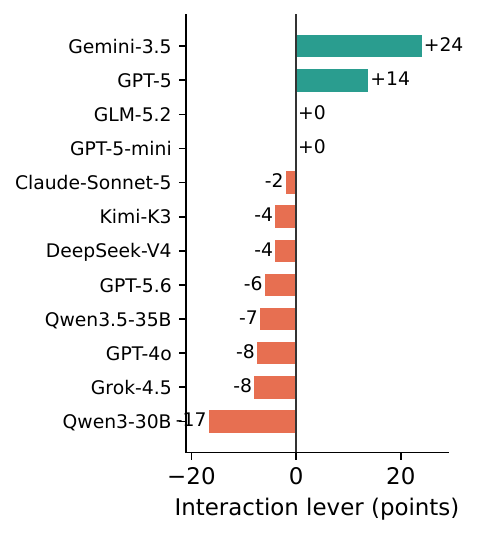}
\caption{\textbf{Interaction is a capability, not a free lever.} Added-interaction
minus plain-search accuracy across the twelve full-scale and cross-vendor models.
Interaction improves accuracy for only GPT-5 and Gemini and is flat or negative for
the other ten, so the ability to convert clarification into a correct answer is rare.}
\label{fig:lever}
\end{figure}

\textbf{Finding 9: Interaction is a capability, not a free lever, it helps
the strong model and \emph{harms} the weak one.} Adding interaction raises
GPT-5 from 6.4\% to 20.2\% (95\% CI [14.5, 26.6]; paired bootstrap $p{<}0.001$),
but \emph{lowers} GPT-4o from 12.1\% to 4.6\% ($p{=}0.002$). Forcing the weaker
model to interact makes it significantly worse than not interacting at all. This
extends \interactcomp's ``interaction stagnation'' into a sharper claim: for
models that cannot use it, interaction is actively harmful. Across all twelve
evaluated models the interaction lever (added-interaction minus plain-search
accuracy) is positive for only GPT-5 and Gemini and flat-or-negative for the other
ten (Figure~\ref{fig:lever}).

\textbf{Finding 10: Model rankings flip with the interaction category.} GPT-4o
\emph{outperforms} GPT-5 under pure retrieval (12.1 vs.\ 6.4) yet is
\emph{dominated} by it under interaction (4.6 vs.\ 20.2). A single-mode benchmark
would mis-rank these models; the interaction category is what separates them, which
is precisely the dimension FinInteract is designed to measure.

\textbf{Finding 11: Retrieval cannot substitute for interaction.} Oracle retrieval
lifts GPT-5 to only 6.4\%: because each disambiguating passage contains
\emph{both} the intended and default values, retrieval surfaces the ambiguity but
does not resolve it. Only a targeted clarification collapses the question to a
unique answer, which interaction (20.2\%) begins to do.

\textbf{Finding 12: The weak model's failure is clarification non-integration, not
inability to ask.} GPT-4o asks readily (IR 99\%, 4.7 questions on average) but
fails to \emph{integrate} the user's responses: 87\% of its interactive errors
are non-committal refusals (e.g.\ ``please specify the exact period'') issued
\emph{after} the simulator has answered. It can recognize ambiguity and ask, but
cannot close the loop and commit, a multi-turn integration gap, not a protocol
artifact (only 2\% produced no answer at all).

\textbf{Finding 13: Models are severely overconfident on ambiguous queries.}
Expected Calibration Error ranges from 0.53 (GPT-5, interact) to 0.95 (GPT-4o,
answer-only); models report 85--95\% confidence while answering 0--20\%
correctly. The remaining failure is not one of category \emph{detection}. Under a
human-validated classifier (agreement 0.75, matching inter-annotator agreement 0.73),
both models almost always ask a question that touches the correct ambiguity category (AC@1
0.86 for GPT-5 and 0.94 for GPT-4o). What they lack is precise \emph{central} targeting,
only 0.06 and 0.04 of first questions make the true category their single central concern,
and, above all, integration (Finding~12). The bottleneck is targeting precision and
commitment rather tha category detection, which motivates the category-conditioned policy of
\S\ref{sec:error}.

\textbf{Finding 14: A smaller model collapses into ill-formed clarification, 
it asks the most but commits the least.} GPT-5-mini scores 0.6/0.0/0.0\% across
the three modes. The cause is not ambiguity-blindness but the opposite: it is the
\emph{most} eager to clarify (interact IR 96\%, 4.6 asks/instance), yet it
issues \emph{open-ended} questions (``which fiscal year? please paste the 2023 and
2022 figures'') rather than the protocol's yes/no questions. Manually inspecting
all 173 interact trajectories, 63\% terminate on an unanswered open-ended
question and 35\% exhaust the 10-round budget still asking; only 1 instance
ever commits a value (also wrong), and just 1.3\% were format-unparseable. Even
in Answer+Search, where no interact channel exists, it asks instead of
answering, capturing \emph{neither} the intended nor the default reading
(0.0/0.0\%). This is an extreme form of Finding~12's integration gap and a
distinct failure mode: clarification without commitment. The collapse is not an
artifact of the yes/no protocol: with a \emph{free-form} simulator that answers its
open-ended questions from $C$ (Appendix~\ref{app:limitations}), GPT-5-mini still resolves
0\% across all 50 instances (zero wrong-to-correct conversions), so the failure
is integrating a received clarification, not the interaction grammar.

\paragraph{Language and category breakdown.} Splitting by language confirms the
benchmark's bilingual-difficulty hypothesis: Chinese instances are consistently
harder than English in the interact mode (GPT-5 18.3\% ZH vs.\ 24.5\% EN;
GPT-4o 0.8\% vs.\ 13.2\%), reflecting more complex corporate-structure naming
and metric conventions in A-share filings. Stratifying by primary category (interact
mode), \texttt{metric\_definition} is the hardest well-populated category (GPT-5
12.7\%, n = 63) and \texttt{entity\_scope} the next (23.4\%, n = 94),
while \texttt{temporal\_scope} is comparatively easy (57.1\%, n = 7). This
inverts the general-domain pattern where entity disambiguation dominates, and
identifies metric-definition ambiguity as the distinctive financial challenge.

\paragraph{Difficulty and headroom.} Per-instance accuracy falls monotonically
with disambiguation entropy $H_0$ (from 10\% at $H_0$ of 1 to under 1\% at
$H_0$ about 3.6), validating $H_0$ as a difficulty proxy; \textbf{no full-scale model
solves 79 of 173 instances (46\%) under free interaction}. These are not broken items:
96\% of them become solvable under the context-oracle, so they are hard specifically at
the elicitation step, not invalid or unanswerable. With the best free-interaction accuracy
at 28.9\% full scale (and 44\% in the cross-vendor pilot), the benchmark is far from
saturated. The supplementary clarification-efficiency
metric mirrors the accuracy ranking in interact mode (DisE$^+$ of 0.125 for GPT-5
vs.\ 0.039 for GPT-4o): GPT-5 not only answers more ambiguous queries correctly
but does so with fewer wasted asks.

\subsection{Human Baseline}
\label{sec:humanbaseline}

To bound the task from above under \emph{realistic} interaction, where the
disambiguating context must be \emph{elicited}, not handed over, we collect a
human baseline on a stratified 50-instance subset (32 EN / 18 ZH, balanced by
category). Two finance-literate annotators (one bilingual) face the same input a
model does in the search regime: the ambiguous question and a retrieved passage
containing both candidate values, position-shuffled and unlabeled. Each may ask
\emph{one} yes/no question, which the experimenter answers truthfully from the
withheld context $C$ (Yes/No/IDK), before committing a final answer. This mirrors
the agent's \texttt{search}to\texttt{interact}to\texttt{answer} loop, so the
human row is directly comparable to the +Interact column; we grade it with the
identical GPT-4o-mini grader.

\begin{figure}[t]
\centering
\includegraphics[width=\linewidth]{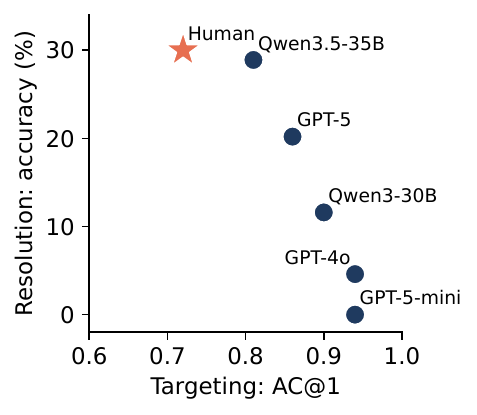}
\caption{\textbf{Targeting does not imply resolution.} Accuracy against AC@1
(touch) in the interact regime, for the five full-scale models and the human baseline
(star). The relationship is inverted: models that ask on the correct category most often
resolve the ambiguity least often, whereas the human asks less precisely yet resolves
more. Asking the right question and answering it are separate abilities.}
\label{fig:humanllm}
\end{figure}

\textbf{Finding 15: Even humans who can ask resolve only 30\%, the task is hard,
not merely mis-specified for models.} Annotators reach 30.0\% accuracy overall
(34.4\% EN, 22.2\% ZH), above every full-scale model (which top out at 28.9\%) but
far from the 93--95\% context-oracle ceiling. The gap between the oracle ceiling and the
human baseline is the cost of \emph{elicitation}: knowing the resolved answer is
near-trivial, but extracting the right bit through a single clarification is hard
even for experts. Across models and the human baseline, targeting and resolution
\emph{dissociate}: the models with the highest AC@1 (GPT-4o and GPT-5-mini at
0.94) post the lowest accuracy, while the human, at a lower touch AC@1 of 0.72
(one focused question rather than a sweep of several categories), posts the highest
(Figure~\ref{fig:humanllm}). The annotators also never targeted the recognition-policy
category either (AC@1 0 on its eight items), so the blind spot is not
model-specific. The asymmetry is sharpest in Chinese, where annotators still capture the
default reading 72.2\% of the time at only 22.2\% accuracy. Chinese instances are
therefore harder to \emph{resolve}, not harder to \emph{recognize}, the
\emph{gui-mu}/\emph{kou-fei} (parent-attributable vs.\ deduction-of-nonrecurring) net-income
distinction is sticky even once correctly surfaced, corroborating the
language-asymmetry caveat of Appendix~\ref{app:limitations} with a mechanism.

\subsection{Skill Dissociation Across Model Scale}

\begin{table}[t]
\centering
\caption{Zero-shot \emph{central}-targeting (central-form AC@1) by base-model scale
(Qwen3, \texttt{+interact}, denominator = instances where the model asked). Under the
validated touch form every scale already reaches about 0.9 on entity and metric, so scale
buys the precision to \emph{center} the question on the right category rather than detection.
Entity central-targeting scales steeply and is largely solved at 32B, while recognition policy
stays 0 at every scale, including the same 32B that targets entity well. Recognition is
underpowered ($n$ at most 9 asked-on) but flat throughout.}
\label{tab:scaledissoc}
\small
\begin{tabular}{lccc}
\toprule
Category & 4B & 14B & 32B \\
\midrule
entity\_scope       & .00 & .00 & \textbf{.68} \\
metric\_definition  & .05 & .00 & .17 \\
recognition\_policy & \textbf{.00} & \textbf{.00} & \textbf{.00} \\
\bottomrule
\end{tabular}
\end{table}

\textbf{Finding 16: the categories are different \emph{kinds} of skill, some
scale-solvable, one a scale-invariant blind spot.} A controlled scale ladder
(Qwen3 4B to 32B, Table~\ref{tab:scaledissoc}) separates the categories
sharply, once we look at \emph{central} targeting. \texttt{recognition\_policy} is a
scale-invariant blind spot. It is never targeted at any scale, in either the touch or the
central form (0.00 at 4B, 14B, and 32B, including the same 32B that targets entity well),
so more parameters buy no recognition-basis targeting at all. \texttt{entity\_scope} and
\texttt{metric\_definition} tell a subtler story. Every scale already \emph{touches} them
(touch AC@1 about 0.9), but the ability to make them the \emph{central} point of the
question emerges only with scale, entity central-targeting rising from 0.00 at 4B and 14B to
0.68 at 32B. Scale therefore does not buy ambiguity detection for these categories, which is
already present, but the precision to center the question on the right category. The dissociation
confirms the taxonomy carves genuine capability boundaries rather than arbitrary strata, and
it pinpoints recognition basis as the category a targeted intervention must address
(\S\ref{sec:coevolve}).

\subsection{Qualitative Case Studies}
\label{sec:casestudies}

Three instances make the failure modes concrete, one for each category where the pattern
is sharpest. Every case pairs a natural analyst question with two defensible readings
whose ground truth is exact and machine-verifiable.

\textbf{Entity scope (resolvable, yet over-elicited).} \emph{``What was Meta
Platforms' operating income?''} is \$46.75B for the total consolidated company and
\$62.87B for the Family of Apps segment, a 34\% gap on the same GAAP metric and
fiscal year. A search-only agent retrieves a passage containing both figures and
commits to one, and single-gold grading credits it whenever that figure happens to
match the key. This is the category models handle best, and the base policy often
resolves it directly (\S\ref{sec:coevolve}), which is precisely why forcing a
clarifying question on entity scope can \emph{lower} accuracy rather than raise it.

\textbf{Recognition policy (the blind spot).} \emph{``What were General Electric's
revenues?''} is \$67.95B under GAAP revenue combining over-time service recognition with
point-in-time product recognition, but \$26.79B counting only revenue recognized at a
point in time, a 2.5 times difference driven entirely by the recognition basis. This is
the category models almost never surface: instead of asking which recognition basis is
meant, the typical clarification reverts to the more salient period or entity question
(\S\ref{sec:skills}), so the ambiguity is never raised and the default is reported as
fact. The category is represented in the network yet not acted on
(\S\ref{sec:mechanistic}). We stress that this category is exploratory (n = 9) and that
an item-level audit (Appendix~\ref{app:limitations}) finds most of its instances to be
ASC~606 timing disaggregations of the kind shown here, where neither value is the total a
non-expert would assume; we therefore treat the pattern as suggestive rather than
quantified.

\textbf{Metric definition (clarification supplies the answer).} \emph{``What was
Workday's operating income growth rate?''} is 126.8\% on a GAAP basis and 26\% on
a non-GAAP basis for the same year-over-year comparison. Here the base model cannot
answer correctly without knowing which definition is intended, so a single on-category
question converts a wrong answer into a right one, the mechanism behind the +45
point training-time gain on this category (\S\ref{sec:coevolve}).

Together the three cases trace the arc of the paper: a category that is resolvable but
over-elicited (entity), a category that is represented yet never acted on (recognition),
and a category where eliciting the definition is the whole battle (metric).

\subsection{Mechanistic Case Study: Where Ambiguity Lives in the Network}
\label{sec:mechanistic}

The behavioral results (\S\ref{sec:exp}) show that frontier models default to the naive
interpretation rather than clarifying. A natural question is whether models internally
\emph{represent} this ambiguity at all. Because this requires access to hidden states, we probe
four open models as a proxy, spanning two size classes and two architectures:
\textbf{Qwen3-4B-Instruct} (dense, 37 layers), \textbf{Qwen3.5-4B} (hybrid linear/full attention,
33 layers), and the mixture-of-experts \textbf{Qwen3-30B-A3B} (49 layers) and
\textbf{Qwen3.5-35B-A3B} (41 layers).

\paragraph{Method.} For each instance we form a within-item contrast: the bare ambiguous
question $Q$ versus $Q$ concatenated with its disambiguating context $C$. We define a per-layer
\emph{ambiguity direction} as the difference in mean final-token activations between the two
conditions, and score any hidden state by its (mean-centered) projection onto that direction;
tracing this across layers gives a depth profile of how strongly each layer separates an
under-specified question from a disambiguated one. Independently, on the two larger models we
train a per-layer linear probe to decode the ambiguity \emph{type} (entity\_scope vs.\
metric\_definition), which, unlike a context-presence detector, cannot be solved trivially.

\begin{figure*}[t]
  \centering
  \includegraphics[width=\textwidth]{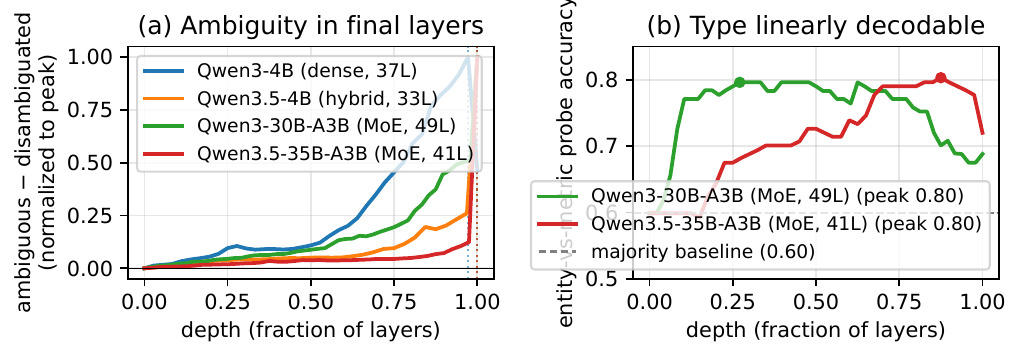}
  \caption{Four-model depth profile of the ambiguity signal (Qwen3 4Bto35B, dense and
  MoE). (a) Ambiguous-disambiguated separation rises to a peak in the final layers of every
  model (dotted verticals mark per-model peaks; normalized for depth and magnitude). (b) The
  ambiguity \emph{type} (entity vs.\ metric) is linearly decodable at 0.80 on both MoE models,
  well above the 0.60 majority baseline, peaking at different depths.}
  \label{fig:crossmodel}
\end{figure*}

\paragraph{Result.} Figure~\ref{fig:crossmodel}(a) overlays the depth profiles (normalized for
differing depth and scale). On all four models the signal is near-zero through most of the stack
and rises sharply to a peak in the \emph{final few layers} (about95--98\% relative depth:
layers 35/37, 32/33, 48/49, 40/41). Figure~\ref{fig:crossmodel}(b) shows that the ambiguity
\emph{type} is linearly decodable at 0.80 accuracy (majority baseline 0.60) on both larger
models, though it becomes available at different depths, early (about27\%) in Qwen3-30B and
late (about85\%) in Qwen3.5-35B. Across four models, two scales, and three architecture
families, the representation is consistent: financial-query ambiguity, and its category, is linearly
encoded, emerging toward the top of the network.

\paragraph{Representation, elicitation, and the cost of interacting.} The two MoE
models we probe are also \emph{evaluated} (Table~\ref{tab:main}): beyond linearly
encoding the category, both \emph{elicit}, and they differ in \emph{how} they ask.
Qwen3-30B interacts on 72\% of instances but loops without committing in Chinese
(76\% of its interactive transcripts never produce a final answer), whereas
Qwen3.5-35B interacts less (36\%) but commits in 95\% of cases and targets the
gold category more often (AC@1 0.40 vs.\ 0.16). With the full retrieval corpus in
place their accuracy is now measured, and it exposes a sharper pattern: plain search
reaches 28.3\%/35.8\%, but \emph{adding} interaction \emph{lowers} accuracy to
11.6\%/28.9\% (-16.8/-6.9 points). Interaction is thus net-negative relative
to search for both open models, as it is for GPT-4o (12.1 to 4.6), and only the
strongest model, GPT-5, converts clarification into a gain (6.4 to 20.2). An
interpretation-oracle control (the resolved interpretation supplied but no evidence)
scores about 0\% for both MoE models, so the residual gap to the oracle ceiling is
an evidence-grounding cost, not a parametric-recall one. These two open models therefore
do act on the represented category to the extent of \emph{asking}, unlike the closed
models, yet asking still fails to convert into a correct answer, which is the same
integration gap seen throughout. A linear-probe replication on the
\emph{base} (non-instruct) checkpoints reproduces the decodability and the early-vs-late depth
split (peaks 0.71/0.74 vs.\ baseline 0.60).

\paragraph{Finding 17: Open models linearly represent both ambiguity and its category, replicating
from 4B to 35B and across dense and MoE architectures.} The representation is type-specific
(entity vs.\ metric decodable at 0.80) and robust to scale and architecture, yet behaviorally
these same models still default rather than clarify. The behavioral gap is thus better read as a
failure to \emph{act} on a represented ambiguity than a failure to register it, though
establishing this causally (e.g.\ via activation steering) remains future work. Concurrent
general-domain work reaches the same behavioral conclusion from the outside, finding that models
identify ambiguity when asked to judge it yet overwhelmingly default to direct
answers~\citep{su2026knowing}; our probe adds that the signal is linearly available in the
representation, and our benchmark supplies the financial setting in which the cost of not acting
on it is exactly measurable.

\paragraph{Scope and caveats.} This is mechanistic evidence on open proxies, not the closed
frontier models benchmarked in \S\ref{sec:exp}; diff-in-means probing and linear decoding
establish \emph{availability}, not causal use; and category-type decoding is supported only for the
two well-populated categories (entity\_scope, metric\_definition). We deliberately do \emph{not} draw
behavioral conclusions from these open models' free-form generations, where surface cues (chain-of-thought
markers, retrieval that surfaces the evidence span) confound naive clarification counts; the
asking-behavior claims come solely from the graded closed-model evaluation in \S\ref{sec:exp}.
A further caveat is that the ambiguity direction is a difference between $Q$ and $Q{+}C$ inputs,
which also differ in length and surface lexicon, so it may partly encode the presence of added
context rather than ambiguity as such. Matched neutral-context controls, a bag-of-words lexical
control, a cross-company split, and causal steering are needed before reading the probe as more
than evidence that category information is \emph{linearly decodable} from the representation.
That the category is \emph{represented but not acted on} frames the bottleneck as an action
problem on an available signal, and motivates the category-conditioned interventions of
\S\ref{sec:error}, which target exactly that signal.

\subsection{Diagnosis and Intervention: Acting on the Ambiguity Category}
\label{sec:error}

Our results converge on a single diagnosis. The benchmark localizes the
bottleneck to \emph{eliciting} the disambiguation: agents answer 93--95\%
correctly once the intended interpretation is supplied
(\S\ref{sec:ceiling}) but far less on their own (a mean of 8\% across the OpenAI
ladder), and the
mechanistic probe shows why, open models \emph{linearly represent} the ambiguity
\emph{category} yet do not act on it (\S\ref{sec:mechanistic}). The remaining gap is
thus an \emph{action} problem on an already-available signal, and the
\emph{category} is its natural unit. We now turn that diagnosis into
intervention: we first catalog how acting fails (a seven-type error taxonomy),
then present two interventions that share one idea, \textbf{condition on the
category}, instantiated at inference time (Category-Aware ReAct, \S\ref{sec:axisreact})
and at training time (category-guided SFT/GRPO, \S\ref{sec:grpo}).

\subsubsection{Error Taxonomy}

Beyond aggregate accuracy, we characterize model failures using a
seven-type error taxonomy (Table~\ref{tab:errors}) derived from
\fininteract's evaluation signals.
Each type is diagnosable from existing metrics without additional annotation.

\begin{table}[h]
\centering
\caption{Error taxonomy E1--E7 with diagnostic signals and improvement directions.}
\label{tab:errors}
\small
\begin{tabular}{p{0.09\columnwidth}p{0.35\columnwidth}p{0.38\columnwidth}}
\toprule
Type & Description & Improvement direction \\
\midrule
E1 & \textbf{Ambiguity blindness}: no interaction, wrong answer
   & Ambiguity pre-check; train on detection \\
E2 & \textbf{Wrong-category clarification}: asked wrong dimension
   & Category classifier; financial taxonomy grounding \\
E3 & \textbf{Generic clarification}: vague ask, no category target
   & Category-conditioned clarification templates \\
E4 & \textbf{Retrieval dominance}: searched but skipped interaction
   & Force interpretation check after retrieval \\
E5 & \textbf{Clarification misuse}: right question, wrong answer
   & State-gated answering; stronger grounding \\
E6 & \textbf{Evidence grounding}: template-oracle correct, agent wrong
   & Improved source-grounded extraction \\
E7 & \textbf{Over-interaction}: correct but low DisE$^+$
   & Penalize via DisE$^+$/AC@1 efficiency signal \\
\bottomrule
\end{tabular}
\end{table}

\subsubsection{Diagnostic Mapping from Evaluation Signals}

The existing evaluation metrics map directly onto error types, enabling
failure attribution without additional annotation:

\begin{itemize}[leftmargin=*,noitemsep]
  \item \textbf{Low IR + low Acc} to E1 (ambiguity blindness dominant)
  \item \textbf{High IR + low AC@1} to E2/E3 (interacts but misdirected)
  \item \textbf{High AC@1 + low Acc} to E5/E6 (right category, wrong grounding)
  \item \textbf{High GenericAskRate} to E3 (lacks financial clarification structure)
  \item \textbf{High WrongCategoryRate} to E2 (confidently misdirected)
  \item \textbf{Always-ask far exceeds Standard ReAct} to E1 dominant; latent capacity exists
  \item \textbf{Template-oracle far exceeds Agent-interact} to ``what to ask'' is the bottleneck
  \item \textbf{Category-oracle far exceeds Agent-interact} to category recognition is the bottleneck
  \item \textbf{Template-oracle still low} to E6 residual; evidence extraction still fails
\end{itemize}

\begin{figure*}[t]
\centering
\includegraphics[width=\textwidth]{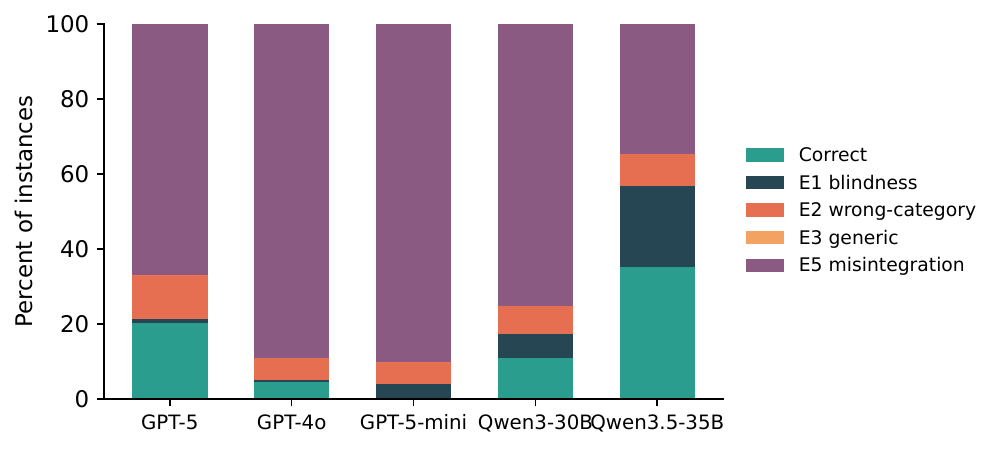}
\caption{\textbf{Error-type distribution} in \texttt{+interact}, attributing every
instance to the taxonomy of Table~\ref{tab:errors} with the corrected category classifier.
\emph{E5 misintegration} (the agent asks, touches the correct category, yet answers wrong)
dominates for every model, while ambiguity blindness (E1), wrong-category (E2), and generic
(E3) asks stay comparatively rare.}
\label{fig:errordist}
\end{figure*}

\textbf{Misintegration, not misdirection, dominates the error budget.} Attributing each
interact-mode outcome to the taxonomy (Figure~\ref{fig:errordist}) shows that \emph{E5}
(right category touched, wrong answer) accounts for the large majority of failures, 67\% for
GPT-5, 89\% for GPT-4o, 90\% for GPT-5-mini, and 75\% for Qwen3-30B, while not-asking
(E1) and off-category asks (E2 and E3) together stay below 15\% for the closed models. The
bottleneck is converting a correctly targeted clarification into the intended answer, not
deciding whether or what to ask. Qwen3.5-35B-A3B is the informative exception, asking less
(E1 21\%) but resolving more of what it attempts (Correct 35\%), so its failures lean
toward under-asking rather than misintegration. This corroborates Findings~7 and~12 across
the full error budget and motivates the state-gated answering of \S\ref{sec:axisreact}.

\subsubsection{Inference-Time Intervention: Category-Aware Clarification ReAct}
\label{sec:axisreact}

The first intervention conditions on the category \emph{without changing the model
weights}. Standard ReAct leaves the agent free to decide whether and what to ask.
We propose \textbf{Category-Aware Clarification ReAct}, a three-component
structured interaction policy that directly addresses E1--E3.

\paragraph{Component 1: Ambiguity Pre-Check.}
Before any action, the agent explicitly decides which of the five ambiguity
categories are underspecified in the question and initializes an interpretation
state $\mathcal{S} = \{$entity, period, metric, basis, vintage$\}$, marking
each field as \emph{resolved} or \emph{unknown}.

\paragraph{Component 2: Category-Conditioned Clarification.}
For each \emph{unknown} field, the agent generates a targeted yes/no question
using category-conditioned templates (one per category) rather than free-form
clarification.
The agent emits at most one clarification per turn, prioritizing the most
information-theoretically valuable unknown field.

\paragraph{Component 3: Interpretation-State-Gated Answering.}
The agent may only emit a final answer once all required fields are resolved
(or the user explicitly declines to clarify).
This prevents the \emph{retrieval dominance} pattern (E4) where the agent
answers immediately after finding a plausible number in a search result.

\paragraph{Expected outcomes.}
Relative to standard Answer+Search+Interact, Category-Aware ReAct should reduce
GenericAskRate (E3), WrongCategoryRate (E2), and IR=0 failures (E1), while
improving AC@1 and ultimately Accuracy.
Template-oracle provides an upper bound: if Category-Aware ReAct approaches
template-oracle performance, it confirms that category recognition is the dominant
remaining bottleneck after the E1--E3 errors are addressed.

\paragraph{Pilot results.}
We evaluate the policy on a GPT-5 backbone over a stratified N = 50 subset run
through the same OpenRouter harness, alongside four matched baselines (Standard ReAct,
Always-Ask, Category-Oracle, Template-Oracle). Category-Aware ReAct reaches 46.0\% accuracy
against the 34.0\% of Standard ReAct in the same run, a +12 point gain, and posts the
best calibration of any condition (ECE 0.241 versus 0.308 for Standard ReAct). Always-Ask
collapses to 10.0\%, confirming that asking indiscriminately is worse than not asking and
that the gain comes from asking \emph{selectively} on the right category. The single-category
diagnostics (AC@1, WrongCategoryRate) are not informative for this policy: because the
Ambiguity Pre-Check deliberately enumerates every candidate category to rule each in or out, an
any-category touch form saturates near 1.0 while the strict central-category form is near 0, so
neither cleanly credits the policy and we omit AC@1 for it in Table~\ref{tab:main}. The
accuracy and calibration gains are the load-bearing result. We report this as a pilot rather
than a full-panel row because it is a single backbone at N = 50, and we expect the effect
size to tighten at full scale. One caveat bears directly on the mechanism: the policy bundles all
three components above, and we do not ablate them here, so part of the +12 may come from the
state-gated commit (Component~3, an integration fix) rather than from better category-conditioned
elicitation (Component~2). A generic-structured clarification baseline that keeps the structure
but drops the named categories, together with a per-component ablation, is the clean control and
remains future work.

\subsubsection{Training-Time Intervention: Category-Guided SFT and GRPO}
\label{sec:grpo}

The per-category skill analysis (\S\ref{sec:skills}) gives this intervention a precise
target: models do not lack clarification ability uniformly, they lack \emph{specific}
category skills (AC@1 of 0 on recognition policy, partial on metric). The question is
whether those missing skills can be installed by training rather than prompting.
The second intervention applies the \emph{same} idea, condition on the
category, but in the weights rather than the prompt, showing the benchmark is not
only diagnosable but \emph{trainable}. Starting from Qwen3-4B-Instruct (4-bit
QLoRA), we run an RLVR ladder, supervised fine-tuning (SFT), KTO, then GRPO, on
the \fininteract training split, with all user simulators, graders, and category
judges served locally (zero API cost). We score each stage with the two
\emph{leak-proof} metrics, AC@1 and interaction rate (Table~\ref{tab:grpo});
accuracy is reported only with the caveat that the training environment's
retrieval returns the disambiguated evidence span, so it is partly leaked and is
\emph{not} a reliable learning signal.

\textbf{The category-guided teacher.} The decisive ingredient is how SFT
demonstrations are generated. An uninformed teacher, even a 72B model, almost
always asks the \emph{obvious} ambiguity (typically the reporting period) and
rarely the subtle gold category (entity scope, recognition policy), producing
about0\% on-category demonstrations. Privately revealing the gold category to the teacher
while generating demonstrations (the hint is \emph{not} stored in the trajectory)
lifts the on-category yield to about79\%, a reusable recipe for any category-structured
clarification task.

\textbf{SFT does the heavy lifting.} The leak-proof signals jump almost entirely
at the supervised stage, AC@1 rises 25 to 69 and interaction 55 to 100
(Figure~\ref{fig:grpo}a), while KTO and GRPO move within run-to-run noise at
n = 51. Training teaches the model to \emph{always interact and target the
correct category}, the behavior \fininteract is built to elicit. Whether this gain
reflects a sharpened internal \emph{representation} of the category or only a rewired
output \emph{policy}, probing the trained model as in \S\ref{sec:mechanistic}, is
a direct test of the diagnosis and a target for future work.

\textbf{True multi-turn GRPO.} The ladder above optimizes a single-turn
approximation. We additionally confirm the full multi-turn episode is trainable
end-to-end: optimizing the 4B policy over entire ReAct episodes (loss masked on
simulator and retrieval tokens) for 15 steps raises mean reward 0.55 to 1.08
while clarifying turns rise 5.7 to 8.6 (Figure~\ref{fig:grpo}b), the category-aware
reward is teaching the policy to ask more. We report this as a learning-signal
demonstration, not a converged model.

\textbf{Caveats.} Results are on a held-out split of n = 51; all judges and the
teacher are local Qwen2.5-32B/72B models (weaker than GPT-4o, and siblings of the
policy); the terminal reward is partly leaked; and GRPO is run for tens of steps,
not to convergence. We therefore frame this as evidence that \fininteract is
\emph{trainable}, a category-guided SFT seed instills category-targeted
clarification, rather than as a state-of-the-art accuracy result.

\subsubsection{When Can the Benchmark Refresh Itself? Three Co-Evolution Probes}
\label{sec:coevolve}

A benchmark that models eventually memorize stops measuring. Because \fininteract
couples a grounded data-generation pipeline (\S\ref{sec:construction}) to a
trainable policy (\S\ref{sec:grpo}), it can in principle \emph{co-evolve}: generate
fresh instances on a category the current model fails, train on them, and repeat, a
data-model loop in the spirit of self-improving
generators~\citep{mae2025}, but \emph{grounded} in real filings rather than the
ungrounded self-play those methods risk. We run three gated rounds, on the scarce,
subordinate category (\texttt{recognition\_policy}), the abundant, salient category
(\texttt{entity\_scope}), and the abundant-but-only-weakly-scale-learnable category
(\texttt{metric\_definition}, \S\ref{sec:skills}), to ask not \emph{whether} the
loop works but \emph{when}.

\begin{table*}[t]
\centering
\caption{Three co-evolution rounds (category-guided SFT on a 4B policy). Targeting
(AC@1) is on a \emph{held-out human} probe never seen in training. The loop
closes on both \emph{abundant} categories and fails only on the \emph{scarce} one; whether
asking \emph{helps accuracy} flips on whether the base could already answer.}
\label{tab:coevolve}
\small
\begin{tabular}{lccc}
\toprule
 & \texttt{entity} & \texttt{metric} & \texttt{recognition} \\
 & \footnotesize abundant & \footnotesize abundant & \footnotesize scarce \\
\midrule
Frontier instances generable        & 99 & 76 & 36 (ceiling) \\
Targeting, base $\to$ evolved        & $.00\to\textbf{.68}$ & $.00\to\textbf{.58}$ & $.00\to.00$ \\
\;\;on held-out \emph{generated}     & .40 & .88 & .00 \\
Interaction rate                     & $40\to100$ & $50\to100$ & $67\to100$ \\
De-leaked accuracy                   & $32.5\to22.5$ & $15\to\textbf{60}$ & $11\to11$ \\
\bottomrule
\end{tabular}
\end{table*}

All three rounds run end-to-end: the ConstructortoVerifier pipeline yields hard
instances (every one passes the adversarial gate: the base fails it answer-only)
and category-guided SFT converges. The outcome splits cleanly by \emph{abundance}, not
by scale-learnability. On both abundant categories, \texttt{entity\_scope} and
\texttt{metric\_definition}, Targeting generalizes to held-out \emph{human} items
(.00 to .68 and .00 to .58, each about4times the +.15 bar) and to held-out
\emph{generated} items, with no catastrophic forgetting. On the scarce
\texttt{recognition\_policy} it does not move (.00 on both held-out sets): the
trained model reverts to the salient period/entity question rather than the
recognition-basis distinction. The contrast rules out a broken pipeline, the
\emph{same} machinery installs entity and metric but not recognition, and, because
metric is only \emph{weakly} scale-learnable yet trains as well as entity, it rules
out scale-learnability as the gate (\S\ref{sec:learnability}).

\textbf{Finding 18: the loop closes on abundant categories and fails on scarce ones;
whether elicitation \emph{helps} is a separate, per-category property.} A self-refreshing
\fininteract is viable wherever the data arm can supply enough frontier
instances, both abundant categories install and generalize; the scarce category
(recognition's 36-instance ceiling from just 30/1,847 primary passages) does
not reliably (Finding~21). But installing on-category elicitation is not automatically beneficial: it
\emph{helps} metric (accuracy 15 to 60, +45: the base cannot answer without the
metric definition, so the clarifying question \emph{supplies} it: 18
wrongtocorrect, 0 regressions) and \emph{hurts} entity (accuracy 32.5 to 22.5,
-10: the base can often answer the entity directly, so forcing a question turns
six correct direct answers into wrong multi-turn ones). Both drove interaction to
100\%; the accuracy sign flips on whether the base could already answer, the
training-time face of the ``interaction can hurt versus plain answering'' effect
seen at inference (\S\ref{sec:ceiling}). The actionable step, which we demonstrate
next, is a \emph{conditional when-to-ask gate}, elicit only where the model cannot
already answer, plus multi-round curriculum and fresh-filing ingestion to lift the
scarce categories above their availability ceiling.

\textbf{Finding 19: the over-elicitation regression is fixable at inference, but
not by a train-time gate.} We test the when-to-ask gate directly on the frozen human
probes (n = 40 per category). A \emph{train-time} gate that mixes no-ask demonstrations
into the category-guided SFT backfires: even 7 of 164 no-ask demonstrations tip the
policy into a bimodal collapse in which it answers directly without retrieving, which
is fatal because in \fininteract almost nothing is answerable from parametric memory
(entity accuracy 32.5 to 7.5, metric 60 to 2.5). An \emph{inference-time} gate
recovers the loss with no extra training. Routing by the base model's own ask
decision, taking the co-evolved asker's answer when the base chose to ask and the
base's direct answer otherwise, lifts entity back to 30.0 from the always-ask
22.5 (matching the base's own 32.5) at a 40\% ask rate rather than 100\%, and
keeps +20 of metric's +45 gain (35.0 versus base 15.0). Over-elicitation is
thus an inference-time routing problem rather than a training one. The residual
metric gap traces to the base's uncertainty signal under-asking where metric needs a
question, so a confidence-calibrated gate that asks when the silent answer would be
\emph{wrong} rather than when the base is merely unsure is the concrete next lever.

\subsubsection{What Gates Co-Evolvability? A Falsification Test}
\label{sec:learnability}

The three rounds let us isolate \emph{which} category property predicts whether the loop
closes. Two candidates are natural, a category's \emph{scale-learnability} (does
zero-shot Targeting rise with model size?) and its \emph{source-abundance} (can the
data arm supply frontier instances?), and we had also hypothesized a
representational version: \emph{learnable iff the category is linearly decodable} from
hidden states. The \texttt{metric\_definition} round was designed to discriminate the
first two; a linear probe tests the third. Table~\ref{tab:learnability} collects the three
candidate predictors against the observed outcome for every category. Both tidy hypotheses fail.

\textbf{Scale-learnability does not gate co-evolution.} \texttt{metric\_definition}
is abundant (498 primary passages) but only \emph{weakly} scale-learnable
(zero-shot AC@1 .05 to .26, slope +0.21, versus entity's +0.77). We
pre-registered the two-factor prediction that it would co-evolve only
\emph{partially}. It did not: metric is a \emph{full} GO (Targeting +0.55, matching
entity's +0.61; Table~\ref{tab:coevolve}). What separates GO from NO-GO is not the
scale-slope but \emph{abundance}, both abundant categories close the loop; only the scarce
category fails.

\textbf{Representation does not gate it either, and is not the bottleneck.} A
per-layer one-vs-rest linear probe decodes every category from base hidden states, and
the pattern \emph{anti}-tracks learnability: \texttt{recognition\_policy}, the
un-learnable, NO-GO category, is the \emph{most} decodable (AUROC 0.93 on a
size-matched sample, above entity/metric at about 0.77); decodability is \emph{flat}
across the 4B to 32B sweep even as behavioral AC@1 climbs
.12 to .89; and category-guided SFT leaves it unchanged. The category is fully
\emph{represented} at every scale, before and after training: the model simply does
not \emph{act} on it, the ``represents-but-doesn't-act'' gap of
\S\ref{sec:mechanistic}, now shown to hold even for the category co-evolution cannot fix.

\begin{table*}[t]
\centering
\caption{What predicts co-evolvability. The observed outcomes track
\emph{abundance/salience}, \emph{not} the AC@1 scale-slope (metric: low slope,
GO) and \emph{not} decodability (recognition: highest AUROC, yet does not
co-evolve). Salience = primary$/$any-candidate passages; decodability = peak
one-vs-rest AUROC (recognition on a size-matched augmented sample; raw n = 9
overfits to 1.0). $^*$metric is a GO but needs about 2 accumulated slices
(about 50 instances); one 30-instance slice teaches it nothing (Finding~21).
$^\ddagger$recognition is \emph{unreliable} rather than a clean fail: held-out human
AC@1 ranges 0--0.44 across training runs on the scarce n = 9 probe.}
\label{tab:learnability}
\small
\begin{tabular}{lrrccc}
\toprule
Category & Primary & Salience & Slope & Decode & Co-evolve \\
\midrule
entity\_scope       & 905 & \textbf{0.92} & $+0.77$ & 0.77 & \textbf{GO} \\
metric\_definition  & 498 & 0.48 & $+0.21$ & 0.77 & \textbf{GO}$^*$ \\
temporal\_scope     & 150 & 0.15 & sat.    & 0.92 & solved \\
filing\_vintage     & 264 & 0.33 & --      & --   & untestable \\
recognition\_policy &  30 & \textbf{0.04} & $+0.00$ & \textbf{0.93} & weak$^{\ddagger}$ \\
\bottomrule
\end{tabular}
\end{table*}

\textbf{Finding 20: co-evolvability is gated by source \emph{abundance}, driven by
\emph{salience}, not by scale-learnability or representational decodability.} What
survives both falsifications is a single upstream cause. \texttt{recognition\_policy}
is a candidate category in 796 passages but \emph{primary} in only 30 (salience
0.04): it is almost always the \emph{subordinate} ambiguity, behind a more salient
entity or period category in the same passage. Low salience produces both the scarcity
that starves the data arm and the salient competitor the model's prior reaches for
instead, so the representation is present but never acted on and cannot be cheaply
trained in. Salient categories behave oppositely (entity, salience 0.92: abundant and
trainable). For a practitioner, \textbf{salience, not scale curves or probe
accuracy, is the cheap, up-front signal} of which categories a co-evolving benchmark can
keep hard and which need fresh human curation. The corollary for evaluation is
sharper: the hardest ambiguity to make models \emph{handle} is not the one they
cannot represent, but the one that is always someone else's more-obvious question.

\begin{figure*}[t]
\centering
\includegraphics[width=\textwidth]{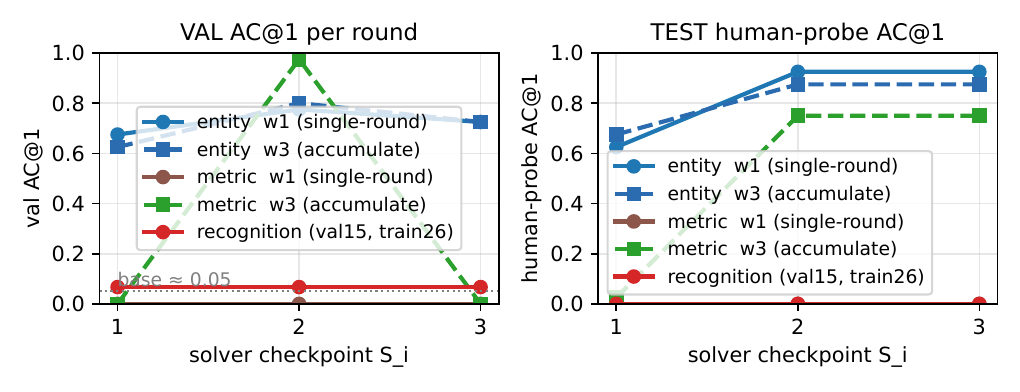}
\caption{Multi-round co-evolution with a validation guard. \emph{Left}: validation
AC@1 (selection metric) per round. \emph{Right}: held-out \emph{human}-probe
AC@1 (never used for selection). Entity learns targeting from a single 30-instance
slice and transfers to humans (0.63 to 0.93); metric needs \emph{accumulation}
(one slice teaches nothing, brown, flat, while two accumulated slices reach
0.75, green); recognition stays at base (about 0.05) even with a fair
26-instance set. The guard preserves human transfer (it rolls back the round-3
candidate, which regresses on validation).}
\label{fig:multiround}
\end{figure*}

\textbf{Finding 21: salience predicts a category's \emph{data-need}, not just a binary
outcome, and a validation-guarded arms race confirms it.} Extending each category to a
multi-round curriculum (Fig.~\ref{fig:multiround}), with checkpoints selected on a
frozen validation set and reported on the held-out human probe, refines the binary
gate into a gradient that tracks salience. \textbf{Entity} (salience 0.92) learns
category-targeting from a \emph{single} 30-instance slice and transfers to human
ambiguity at 0.88--0.93. \textbf{Metric} (0.48) needs \emph{accumulation}: one
slice yields 0.0, two accumulated slices reach 0.75, so its earlier single-round
GO used about 60 instances precisely because that is roughly what the category requires.
\textbf{Recognition} (0.04) is \emph{unreliable}: across training runs its human
AC@1 ranges only 0--0.44 on the scarce n = 9 probe, weak and inconsistent,
not a clean zero. Scoring the same runs as an \emph{Elo arms race} between the solver
and a hard-example miner (win $=$ AC@1, which, unlike recall-walled resolution, %
varies across checkpoints) makes the instrument non-degenerate: entity is
\emph{solver-runaway} (solver Elo 1000 to 1551, outpacing a corpus-bounded miner
that cannot synthesize adversarially harder targeting), while recognition
\emph{stalls} (flat). Two guarded rounds peak; the third overfits and is rolled back.
The practical reading: salience is an up-front estimate not just of \emph{whether} an
category co-evolves but of \emph{how much} data it needs, and one well-targeted round,
guarded on held-out validation, is close to the sweet spot.

\section{Extended Related Work}
\label{app:related}

This appendix expands the related work summarized in \S\ref{sec:related}.

\begin{table*}[t]
\centering
\setlength{\tabcolsep}{5pt}
\caption{Comparison of financial QA benchmarks. A check mark denotes a
supported feature; -- denotes not applicable or not evaluated.}
\label{tab:comparison}
\small
\begin{tabular}{p{3.1cm}ccccccc}
\toprule
Benchmark & Scale & Ambig. & Interact. & Multi-turn & Biling. & Evidence & Grounding \\
\midrule
FinanceBench \newline \citep{islam2023financebench}
  & 150 & -- & -- & -- & -- & human & SEC \\
FinQA \newline \citep{chen2021finqa}
  & 8,281 & -- & -- & -- & -- & human & 10-K/Q \\
DocFinQA \newline \citep{reddy2024docfinqa}
  & 7,437 & -- & -- & -- & -- & human & 10-K/Q \\
BizBench \newline \citep{krumdick2024bizbench}
  & multi-task & -- & -- & -- & -- & auto & mixed \\
FinBen \newline \citep{xie2024finben}
  & multi-task & partial & -- & -- & partial & auto & mixed \\
FinanceQA \newline \citep{mateega2025financeqa}
  & multi-task & -- & -- & -- & -- & expert & SEC \\
Finance Agent Bench. \newline \citep{bigeard2025financeagent}
  & 537 & -- & -- & -- & -- & expert & SEC \\
FinAgentBench \newline \citep{choi2025finagentbench}
  & 26K & -- & -- & -- & -- & auto & SEC \\
FinEval \newline \citep{guo2023fineval}
  & 8,351 & -- & -- & -- & ZH & expert & exams \\
\midrule
\fininteract (ours)
  & 173 & \checkmark & \checkmark & \checkmark & EN+ZH & LLM+human & EDGAR+CNINFO \\
\bottomrule
\end{tabular}
\end{table*}

\paragraph{Financial QA Benchmarks.}
Table~\ref{tab:comparison} compares \fininteract to major financial QA
benchmarks.
FinanceBench~\citep{islam2023financebench},
FinQA~\citep{chen2021finqa},
DocFinQA~\citep{reddy2024docfinqa},
BizBench~\citep{krumdick2024bizbench},
FinBen~\citep{xie2024finben}, and
FinanceQA~\citep{mateega2025financeqa}
all enforce single-gold-answer conventions and evaluate models in a single turn;
FinanceQA further shows that frontier models fail about60\% of realistic
analyst tasks, but attributes this to numerical rigor rather than ambiguity.
FinEval~\citep{guo2023fineval}, a large Chinese financial-knowledge benchmark,
incidentally identifies an ``ambiguity handling weakness'' error type in its
agent tasks but does not isolate or control it. Recent \emph{agentic} financial
benchmarks raise task realism, Finance Agent Benchmark~\citep{bigeard2025financeagent}
(537 expert research tasks, where the best model reaches only 46.8\%) and
FinAgentBench~\citep{choi2025finagentbench} (agentic retrieval over EDGAR), but both
remain single-turn and single-gold, evaluating analysis and retrieval rather than
whether an agent recognizes that a query is under-specified. \fininteract is the
first to make query ambiguity the central, controlled construct, via paired
default/intended interpretations and multi-turn interaction, in the financial
domain.

\paragraph{Ambiguous QA.}
AmbigQA~\citep{min2020ambigqa},
ASQA~\citep{stelmakh2022asqa},
CondAmbigQA~\citep{li2025condambigqa}, and
CLAM~\citep{kuhn2022clam}
have matured in the general (Wikipedia) domain.
More recent benchmarks target the agent's \emph{own} ability to ask:
CLAMBER~\citep{zhang2024clamber} evaluates whether models identify and clarify
ambiguous information needs, and QuestBench~\citep{li2025questbench} tests whether
they can ask the minimal necessary question in reasoning tasks. Closest in spirit to our
diagnosis, \citet{su2026knowing} show in the general domain that models recognize ambiguity when
asked to judge it yet rarely ask about it in ordinary question answering, and that retrieved
context makes them ask \emph{less}, a result our retrieval-augmented rows reproduce in finance.
These works are general-domain and judge a single clarifying step; \fininteract adds \emph{per-category}
targeting (AC@1), programmatically verifiable financial grounding, and, via the
co-evolution rounds (\S\ref{sec:coevolve}), an analysis of \emph{when} asking helps
versus hurts. We adapt this line's formalism to finance, replacing human-judged
ambiguity labels with programmatically verifiable XBRL-grounded evidence pairs.

\paragraph{Clarifying Questions.}
Asking clarifying questions is a long-standing goal in information retrieval and
dialogue. Rao and Daum\'e III~\citep{rao2018clarification} rank clarification
questions by expected value of perfect information; ClariQ~\citep{aliannejadi2020clariq}
established a shared task for generating clarifying questions in open-domain dialogue;
and the mixed-initiative paradigm is surveyed in Conversational Information
Seeking~\citep{zamani2023cis}. \fininteract instantiates this lineage with a
leak-proof, per-category measure of \emph{which} clarification an agent should issue and a
verifiable outcome for whether issuing it resolves the query.

\paragraph{Interactive Search Agents.}
\interactcomp~\citep{interactcomp2026} is the direct methodological ancestor.
We extend its general-domain setting to finance with a domain-specific
ambiguity taxonomy and bilingual evaluation.
BrowseComp~\citep{wei2025browsecomp} tracks longitudinal retrieval trends;
we test whether the interaction-capability gap identified there persists in
finance.


\section{Extended Discussion}
\label{app:discussion}

This appendix expands the discussion points summarized in \S\ref{sec:discussion}.

\paragraph{Ambiguity categories are different learning problems.} The per-category dissociation shows that some
clarification abilities are solved by scale while others, in particular the recognition basis, resist
it and are only unreliably teachable. A single aggregate interaction score therefore hides
where the real difficulty sits. Future benchmarks and training curricula should treat each
category as its own learning problem and budget data accordingly, which our salience estimate
begins to quantify.

\paragraph{Elicitation is a trainable skill that current models lack.} Our central result
is that models fail to resolve ambiguity not for want of knowledge but for want of asking
the right question (RQ2). This reframes the problem for future work. Progress will come less
from larger backbones, which barely move interaction accuracy, and more from policies and
objectives that reward asking on the correct category at the right time. The Category-Aware ReAct
policy and category-guided fine-tuning (RQ3) are first steps, and both leave a large gap to the
oracle ceiling, so reward shaping on the ambiguity category and calibrated deferral are natural
next directions.

\paragraph{Ambiguity persists in human-AI collaboration.} Because models confidently return
the default reading, the safe behavior is often to ask rather than to answer. This motivates
evaluation and interfaces that credit well-timed clarification and calibrated abstention
rather than raw commitment, and it connects to mixed-initiative retrieval where the cost of
a confident wrong answer is high.

\paragraph{The paired construction should transfer to further categories, languages, and high-stakes domains.} Our co-evolution probes suggest a
benchmark can extend itself on categories whose evidence is abundant, guarded by held-out
validation, which offers a path to keep the instrument fresh as models improve. Natural
extensions include more ambiguity categories, additional languages and filing regimes, and
transfer of the paired default and intended construction to other high-stakes domains where
a single gold answer hides genuine ambiguity.


\section{Extended Limitations}
\label{app:limitations}

This appendix expands each limitation summarized in \limref.

\textbf{Scale.} At 173 instances FinInteract is small (comparable to
FinanceBench's 150~\citep{islam2023financebench}); we prioritize per-instance
construction rigor (every item is adversarially verified and R14-discriminating)
over raw size, and Appendix~\ref{app:cost} shows the pipeline scales at
\$0.31/instance.

\textbf{Language balance.} The English split (53) is smaller
than the Chinese split (120); per-language results (\S\ref{sec:exp}) should be
read with this asymmetry in mind, and the EN portion is a target for growth.

\textbf{Category coverage and recognition-policy validity.} The taxonomy is intentionally
five-category, but \texttt{filing\_vintage} is unrepresented and \texttt{temporal\_scope}/%
\texttt{recognition\_policy} are sparse (Table~\ref{tab:stats}): viable
structured ambiguity for these categories is rare in public filings (Appendix~\ref{app:cost}).
We therefore treat entity scope (n = 94) and metric definition (n = 63) as the two
well-powered categories and the other three as exploratory. An item-level audit of the nine
\texttt{recognition\_policy} instances is a further caution: most are ASC~606 timing
disaggregations (revenue recognized point-in-time versus over-time), where neither value is
the total a non-expert would assume, and one pairs total GAAP revenue against an operating
cash-flow proxy that is not a defensible reading of ``revenue.'' Even our human annotators
never targeted the category (AC@1 0 on its items). We therefore report the
recognition-policy blind spot as a preliminary observation on a small, imperfect sample, not
a headline result, and recommend rebuilding this category with analyst-audited pairs before
drawing quantitative conclusions.

\textbf{Model coverage.} We report the OpenAI tier ladder (3 models) and two open
MoE models at full scale (n = 173); a seven-model, six-vendor panel of current
frontier models is additionally evaluated in a preliminary pilot (N = 50/mode,
Table~\ref{tab:main}, panel (b)), which reproduces the central finding but at reduced
statistical resolution, and a full-scale multi-vendor evaluation remains in progress.
The human baseline (N = 50, a single yes/no clarification) is a constrained reference
rather than an upper bound, since models given up to ten rounds can exceed it, and it is
collected from two annotators on a subset. A larger expert panel remains future work.

\textbf{Grader and simulator.} Correctness and
category-targeting are judged by GPT-4o-mini and the user is simulated by GPT-5. We validate
these against humans on a stratified, blindly double-annotated sample (five annotators).
The correctness grader agrees with human consensus at Cohen $\kappa{=}0.85$ (raw agreement
0.92), exceeding inter-annotator agreement (0.86), so accuracy figures hold under human
labels. The AC@1 classifier initially agreed only weakly (0.53). A single-label design
over-assigned \texttt{temporal\_scope} to compound questions that also name an entity or
metric. The corrected multi-label \emph{touch} classifier reaches 0.75 agreement, matching
inter-annotator agreement (0.73), and the AC@1 values reported here use it. Because
this shows models mostly do ask on-category, the corrected reading is that the gap lies in
central targeting and integration rather than detection. A strict central-category variant is
discriminative but not yet human-validated, so we report it only as a supplementary
decomposition. Interaction results are also robust to the simulator: on a
50-instance subset, standard +Interact accuracy is 34/28/26\% under the LLM, deterministic,
and 15\%-noisy simulators. Crucially, the yes/no action grammar does not manufacture the
weak-model collapse: with a \emph{free-form} simulator that answers open-ended clarifying
questions from $C$ (for example ``which fiscal year?'') rather than only yes/no, GPT-5-mini
stays at 0\% across all 50 instances (identical to the yes/no simulator, zero
wrong-to-correct conversions, three-round budget). Its failure is therefore a
post-clarification integration failure, not an artifact of the interaction protocol.

\textbf{Construction circularity and synthetic intent.} The constructor, adversarial verifier, and user
simulator are all GPT-5-family models, and the verifier removes instances the GPT-5 family
answers correctly without $C$. This depresses GPT-5-family answer-only accuracy (0.6\%) by
construction relative to the non-OpenAI vendors the filter does not target (mean 10.0\%
answer-only across six vendors), so cross-vendor answer-only comparisons are not fully
apples-to-apples and the GPT-5-family floor is a filter-induced lower bound. The elicitation
gap itself is unaffected: the filter targets answer-only solvability, not interactive
resolution, and every vendor still resolves far below its oracle ceiling. Verifying a
subsample with a non-OpenAI adversarial ensemble is a natural robustness check for future work.
A related construct-validity caveat is that questions are reverse-constructed from filings
rather than drawn from real user queries, and the ``intended'' interpretation is a
constructor-assigned hidden context. The benchmark therefore measures a model's ability to
recover a \emph{specified} interpretation, a necessary and controllable proxy for, but not
identical to, disambiguating the intent of a real analyst.

\textbf{Contamination.}
Instances are freshly constructed from FY2022--2025 structured facts and
answer-only accuracy is about0.6\% across all sources, but we do not claim
absolute contamination safety for the DocFinQA-derived minority.

\textbf{All instances are ambiguous by construction.} Every one of the 173 items is built to
require clarification, so a model can succeed by assuming ambiguity is always present.
FinInteract therefore measures \emph{what} to ask and \emph{how} to integrate the answer, not a
calibrated \emph{when}-to-ask policy: it does not score false-positive clarification, over-asking
on already-specified questions, or the correct choice to answer directly when a mild ambiguity
does not change the value. Adding matched unambiguous and answer-invariant controls is the single
most important extension for turning the benchmark into a test of calibrated interaction, and we
accordingly frame our claims around asking-and-integrating rather than when-to-ask.

\textbf{Intervention evidence is preliminary and not component-attributed.} The Category-Aware
ReAct gain (34 to 46\%) is a single GPT-5 backbone on N = 50 (about six additional items),
without a significance test, and its matched-baseline subset scores above the full-scale GPT-5
row, so the effect must be confirmed at full scale with confidence intervals. Just as important,
the policy bundles three components (ambiguity pre-check, category-conditioned questioning, and
interpretation-state-gated answering) that we do not ablate: the improvement may stem
substantially from the state-gated commit, an \emph{integration} fix, rather than from better
category \emph{elicitation}, so the pilot does not by itself isolate the taxonomy's mechanistic
role. A generic-structured clarification baseline and a per-component ablation are the needed
controls. The training-time results are likewise early, resting on a small held-out probe, gains
dominated by supervised fine-tuning with only a short reinforcement phase, an optimistic
oracle-retrieval reference, and interaction driven to near-total compliance rather than a learned
when-to-ask policy.

\textbf{A single protocol confounds ability with interface adherence.} All vendors share one
ReAct prompt and yes/no action grammar. GPT-5-mini's 0\% is partly a protocol-adherence failure
(it keeps asking open-ended questions, exhausts the round budget, and rarely commits a final
answer) rather than proof of zero financial-ambiguity ability. A free-form simulator rules out
the yes/no grammar as the cause (the collapse persists), but tool schema, few-shot examples,
constrained decoding, and model-specific adapters all shape protocol compliance, so a fixed
cross-vendor prompt is a conservative rather than a maximally fair comparison; per-model interface
tuning is future work.

\textbf{Scope.} The paper spans a benchmark, the single-gold illusion, the taxonomy, and several
exploratory studies (representation probing, inference and training interventions, co-evolution,
and salience). The three main-text findings are the load-bearing contribution and the rest are
explicitly exploratory and appendix-scoped; the two highest-value additions they point to,
matched ambiguous/unambiguous evaluation and a component ablation of the intervention, are our
priority follow-ups.

\section{Curation Cost to Replicate the Benchmark}
\label{app:cost}

A practical concern for any LLM-assisted benchmark is the cost to reproduce or
scale it. We instrument the construction pipeline (\S\ref{sec:construction}) and
report measured API costs. All figures use list prices at construction time
(per~1M tokens: GPT-5 \$1.25 in / \$10.00 out; GPT-5-mini \$0.25 / \$2.00;
GPT-4o-mini \$0.15 / \$0.60) and count API fees only, they exclude the one-time
filing-collection compute and the human spot-check labor (10--20\% of
accepted instances).

\paragraph{The constructor, not the verifier, is the yield bottleneck.}
Re-running the pipeline over a held-out sample of candidate passages reproduces
a sharp funnel (Table~\ref{tab:cost}). Only 11.5\% of keyword-tagged candidate
passages survive the constructor, which rejects the rest as containing
\emph{no viable ambiguity} for the target category; but of those that survive,
98.5\% pass the 10-trial adversarial verifier. In other words, the
construction prompt is strict and the verifier rarely needs to fire, the cost
of the benchmark is dominated by constructor calls spent on candidate passages
that turn out not to support a genuine ambiguity, not by verification.

\begin{table}[!htbp]
\centering
\caption{Measured curation cost. Unit costs are from instrumented pipeline runs
(constructor-only n = 20; full pipeline n = 6); the acceptance funnel is from
the full construction run (n = 1127 candidate passages). A full pipeline pass
is one GPT-5 constructor call plus ten verifier trials (five GPT-5, five
GPT-5-mini) with GPT-4o-mini grading.}
\label{tab:cost}
\small
\begin{tabular}{lr}
\toprule
Quantity & Value \\
\midrule
Constructor viability (viable / candidate)      & $11.5\%$ \\
Verifier acceptance (accepted / viable)         & $98.5\%$ \\
Overall acceptance (accepted / candidate)       & $11.3\%$ \\
\midrule
Cost per rejected candidate (constructor only)  & \$0.020 \\
Cost per viable passage (full pipeline)         & \$0.155 \\
Blended cost per candidate passage              & \$0.035 \\
\textbf{Cost per accepted instance}             & \textbf{\$0.31} \\
\bottomrule
\end{tabular}
\end{table}

\paragraph{Projected cost to scale.} At the measured blended rate, producing $N$
accepted instances requires about $N/0.113$ candidate passages and the
following API spend: 173 instances (the current release) about \$54;
500 instances about \$155; 1000 instances about \$309. The
benchmark is therefore inexpensive to scale in API terms, an order of
\$300 takes it to 1000 instances, and the true constraint is the supply of
candidate passages that contain structured, verifiable ambiguity. This favors
\emph{better passage pre-selection} (raising the 11.5\% viability rate) over
brute-force compute, and explains why the rare categories (e.g.\ filing\_vintage),
whose candidate passages almost never survive the constructor, are the expensive
ones to grow.

\section{Construction and Validation Details}
\label{app:construction}

This appendix expands the data sources, curation pipeline, and validation summarized in
\S\ref{sec:methodology}.

\begin{figure*}[t]
\centering
\includegraphics[width=\textwidth]{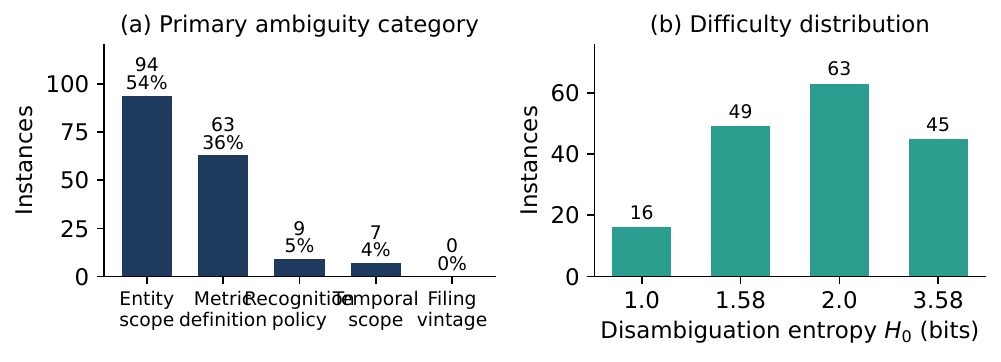}
\caption{\textbf{\fininteract distributions.} \emph{(a)}~Primary ambiguity category: the
corpus is entity/metric-dominant (54\% / 36\%), with the three subordinate categories
sparse, mirroring how often each category is the \emph{primary} ambiguity in public
filings (Sec.~\ref{sec:task}). \emph{(b)}~Difficulty by disambiguation entropy
$H_0=\log_2 k$: instances span one to 3.58 bits, most at one or two bits with a
substantial multi-category tail at 3.58 bits (45 instances).}
\label{fig:datastats}
\end{figure*}

\paragraph{Data sources.}
Figure~\ref{fig:datastats} summarizes the resulting category and difficulty distributions.
\textbf{English (EN, about80\% of pool).} \textit{DocFinQA}, 780 long-context passages extracted
from 10-K and 10-Q filings, with verified candidate answers from annotated QA pairs.
\textit{EDGAR (XBRL-derived)}, 111 passages built from machine-readable XBRL facts for S\&P~500 and
Russell~1000 companies (FY2024--2025). \textbf{Chinese (ZH, about20\% of pool).}
\textit{CNINFO/akshare}, 237 passages derived from A-share annual reports (SSE~50 + CSI~300) via
structured financial-statement APIs, in three passage types: (1) metric definition, \emph{kou-fei}
vs.\ \emph{gui-mu} (non-recurring-excluded profit vs.\ parent-attributable profit); (2) entity
scope, \emph{gui-mu} vs.\ \emph{jing-lirun} (parent-attributable vs.\ fully consolidated including
minority interest); (3) temporal scope, year-over-year revenue growth. ZH instances are expected to
be harder for current models due to sparser training data on A-share filings and more complex
corporate-structure nomenclature (e.g., A/H-share distinctions, \emph{jituan} (group) vs.\
\emph{zi-gongsi} (subsidiary) scope).

\paragraph{Three-role construction pipeline.}
\textbf{Step 1: Constructor (GPT-5).} A single GPT-5 call generates the full $(Q, C, A)$ triple
given a filing passage, verified candidate answer, and mandatory primary category. The output JSON
includes \texttt{question}, \texttt{context}, \texttt{answer}, \texttt{default\_answer},
\texttt{intended\_evidence\_span}, \texttt{default\_evidence\_span}, \texttt{intended\_interpretation},
\texttt{default\_interpretation}, and \texttt{axes\_exercised}, with category-specific instructions
injected per primary category. \textbf{Step 2: Pre-verifier sanity checks.} Fourteen automated
code-level rules (Table~\ref{tab:qc}) are applied without API calls, and any failure immediately
rejects the passage. \textbf{Step 3: Adversarial verifier.} Ten verifier calls run in parallel
(five GPT-5-mini and five GPT-5); each answers $Q$ without $C$ and states its assumed
interpretation, and an instance is rejected if at least 2 of the 10 both produce $A$ and align with
the intended interpretation. This two-condition criterion prevents false rejections from lucky
guesses.

\paragraph{Category diversity enforcement.}
We enforce target category shares through four mechanisms: (1) a priority function with a 3 times
penalty on over-quota categories; (2) hard exclusion at 2 times quota; (3) a rarity-adjusted
initial sort (passages sorted by
$\sum_\text{categories}(\text{target\_share}/\text{pool\_frequency})$); and (4) a dynamic re-sort
every 10 instances.

\paragraph{Annotator recruitment, payment, and review status.}
Annotators are finance professionals recruited individually by cold email, with no
institutional collaboration or prior relationship to the authors, and no party in a position
of authority over them; participation was voluntary and could be ended at any time. They were
paid an agreed rate of USD 50 per hour of annotation, set to be commensurate with professional
freelance rates for comparable financial-analysis work. Because the study engaged a small
number of professional annotators on this basis, collected no personal data about them, and
had them label only public corporate filings, we did not seek an ethics review board
determination; no institutional review framework applied to the arrangement. Readers planning
a larger or differently constituted annotation effort should confirm the requirements of their
own institution.

\paragraph{Human validation protocol.}
Two finance-literate annotators (one bilingual, for the Chinese split) independently review
instances on an eight-question protocol: H1 \emph{Is $Q$ ambiguous without $C$?} H2 \emph{Is the
default interpretation plausible for a non-expert?} H3 \emph{Does $C$ uniquely identify the intended
interpretation?} H4 \emph{Does $C$ avoid directly stating the answer?} H5 \emph{Is $A$ correct under
the intended interpretation?} H6 \emph{Is $A_d$ correct under the default interpretation?} H7
\emph{Is the primary category label correct?} H8 \emph{What yes/no question would a financial analyst
naturally ask to resolve the ambiguity?} (free text, used to compute human AC@1). Acceptance
requires H1--H6 to hold after adjudication by the primary author. On a stratified 60-instance sample
(by language and category), per-item raw agreement is 0.82--0.97. Because the labels are heavily
skewed toward ``yes'' ($>85\%$ per item), Cohen's $\kappa$ is degenerate under the
Feinstein--Cicchetti prevalence paradox, so we report Gwet's AC1 (0.85 average, range
0.78--0.97; 1.00 for H7). The single consensus-rejected instance (consensus rejection
1.7\%) was a category over-reach, a revenue-recognition-timing pairing whose plausible default is
the firm's total revenue rather than a disaggregation component. It is answerable only under
context-oracle, where removing it shifts the ceiling by $<\!0.05$pp (93--95\% unchanged).

\section{Construction Quality-Control Rules}
\label{app:qc}

The pre-verifier applies fourteen automated code-level rules before any API call (Table~\ref{tab:qc}), rejecting a candidate the moment any rule fails.

\begin{table}[h]
\centering
\caption{Pre-verifier sanity checks (14 rules).}
\label{tab:qc}
\small
\begin{tabular}{cp{0.72\columnwidth}}
\toprule
Rule & Description \\
\midrule
R1  & $Q$ must not be answerable yes/no \\
R2  & $Q$ must not contain disambiguating terms or inline dates \\
R3  & $A$ taken verbatim from the source candidate value (not invented) \\
R4  & $Q$ must name the company \\
R5  & $Q$ must ask about a substantive financial metric \\
R6  & Answer type consistent with question type \\
R7  & Context $C$ must not contain the answer value \\
R8  & Intended and default interpretations must differ \\
R9  & All category labels must be valid vocabulary items \\
R10 & Answer must be non-trivial (not empty or ``0'') \\
R11 & \texttt{default\_answer} present and differs from \texttt{answer} \\
R12 & Both evidence spans non-empty \\
R13 & Evidence spans must be meaningfully distinct ($\leq$85\% overlap) \\
R14 & Intended and default \emph{answers} differ beyond grader tolerance \\
\bottomrule
\end{tabular}
\end{table}

\section{Reproducibility}
\label{app:repro}

\textbf{Models and decoding.} Closed models are accessed via their official APIs
(evaluated 2026): \texttt{gpt-5}, \texttt{gpt-4o}, \texttt{gpt-5-mini}. Reasoning
models (gpt-5 family) use default temperature with a generous
\texttt{max\_completion\_tokens} headroom (+4096 above the output budget) so
internal reasoning does not truncate the answer; other models use temperature
0 for the agent and grader. \textbf{User simulator.} GPT-5 at temperature 1.0,
constrained to \{Yes, No, I don't know\} and conditioned on the disambiguating
context $C$. \textbf{Grader.} GPT-4o-mini at temperature 0, binary correctness
with finance tolerance (plus or minus 1\% numeric, entity/ticker equivalence, currency
normalization; fiscal-year mismatch = wrong). \textbf{Protocol.} ReAct with
\texttt{search}/\texttt{interact}/\texttt{answer} actions, max 10 rounds; the
forced-interaction ablation requires n = 4 asks before answering; the
context-oracle ceiling supplies $C$ plus both evidence spans (position-shuffled).
\textbf{Statistics.} Accuracy CIs are 95\% bootstrap over instances (B = 2000);
mode transitions use a two-sided paired bootstrap (B = 5000).
\textbf{Compute.} Evaluation is API-based and its cost is reported in dollars rather than
GPU hours (Appendix~\ref{app:cost}); the only local compute is the open-weight probing and
the 4-bit QLoRA training of the 4B policy, together with the locally served simulators,
graders, and category judges, run on \GPUSPEC. Local compute excluding training took approximately 6 GPU hours. All evaluation,
construction, and analysis scripts are released.

\section{Datasheet and Licensing}
\label{app:datasheet}

\textbf{Provenance.} English instances derive from public SEC EDGAR filings
(10-K, FY2022--2025) and a small DocFinQA-sourced set; Chinese instances derive
from public CSRC annual reports retrieved via CNINFO/akshare. All source
documents are public regulatory filings; no proprietary, paywalled, or
personal data is used. \textbf{Schema.} Each instance is a JSON record with
fields: \texttt{instance\_id}, \texttt{language}, \texttt{source}, \texttt{ticker},
\texttt{company}, \texttt{filing\_type}, \texttt{filing\_date}, \texttt{question},
\texttt{context}~($C$), \texttt{answer}, \texttt{default\_answer},
\texttt{intended\_evidence\_span}, \texttt{default\_evidence\_span},
\texttt{intended\_interpretation},
\texttt{default\_interpretation}
(\{entity, period, metric, basis\}), \texttt{axes}, \texttt{n\_axes},
\texttt{h0}, and \texttt{qc} (per-rule pass flags + verifier blind-solve rate).
\textbf{Intended use.} Evaluating whether interactive search agents recognize and
resolve query ambiguity; not a training corpus and not financial advice.
\textbf{License and release.} The benchmark, construction pipeline, and evaluation
code are released for research use; redistribution of source filings follows the
respective public-domain (EDGAR) and regulatory-disclosure (CSRC) terms.
\textbf{Maintenance.} The frozen v1 release is versioned; corrections and the
planned EN-split expansion will be issued as numbered releases.

\section*{Disclosure of Generative AI Use}

Generative AI tools were used in this work in two distinct roles. First, as
\emph{components of the research artifact itself}: large language models act as the
adversarial constructor, the user simulator, and the automated grader described in
Section~\ref{sec:methodology} and Section~\ref{sec:eval}, and every such use is
documented in the paper and the accompanying code with the specific model, version,
and prompt. Second, generative AI tools were used to assist with \emph{manuscript and
code preparation}, in four respects: prose editing and restructuring, including
condensing and reordering material we had already written; \LaTeX{} formatting,
cross-reference repair, and preparation of the alternate-format build; consistency
checking of the manuscript against our own result files, which surfaced mismatches we
then corrected; and auxiliary tooling for packaging the supplementary archives and for
normalizing in-text notation. One literature search was AI-assisted and led us to a
concurrent work that we read, verified, and cite.
All research questions, experimental design, data construction decisions, experiments,
analyses, and scientific claims are the authors' own. No reported result, table, or
figure was produced by an AI assistant; every number derives from our own experimental
runs. The authors have independently verified every result, table,
figure, and citation reported here, and take full responsibility for the
entire content of this paper, in accordance with \ethicscode.

\end{document}